\documentclass[10pt,twocolumn,letterpaper]{article}

\usepackage[pagenumbers]{cvpr} 

\usepackage{graphicx}
\usepackage{amsmath}
\usepackage{amssymb}
\usepackage{booktabs}
\usepackage{caption}
\usepackage{multirow}

\usepackage{amssymb}
\usepackage{threeparttable}

\def\BibTex{{\rm B\kern-.05em{\sc i\kern-.025em b}\kern-.08em
    T\kern-.1667em\lower.7ex\hbox{E}\kern-.125emx}}
\newcommand{\tabincell}[2]{\begin{tabular}{@{}#1@{}}#2\end{tabular}}

\usepackage{array}
\makeatletter
\newcommand{\thickhline}{%
    \noalign {\ifnum 0=`}\fi \hrule height 0.8pt
    \futurelet \reserved@a \@xhline
}

\makeatletter
\let\NAT@parse\undefined
\makeatother
\usepackage[colorlinks,
            linkcolor=red,       
            anchorcolor=blue,  
            citecolor=green,        
            ]{hyperref}

\usepackage[capitalize]{cleveref}
\crefname{section}{Sec.}{Secs.}
\Crefname{section}{Section}{Sections}
\Crefname{table}{Table}{Tables}
\crefname{table}{Tab.}{Tabs.}

\def\cvprPaperID{*****} 
\def\confName{CVPR}
\def\confYear{2024}

\begin{document}


\title{Brain-Conditional Multimodal Synthesis: A Survey and Taxonomy}

\author{Weijian Mai\textsuperscript{1},  Jian Zhang\textsuperscript{1}, Pengfei Fang\textsuperscript{2}, Zhijun Zhang\textsuperscript{1*}\\
\textsuperscript{1}South China University of Technology \\
\textsuperscript{2}Southeast University\\
{\tt\small \textit{auzjzhang@scut.edu.cn}}
\thanks{Corresponding author}
}
\maketitle

\begin{abstract}
   In the era of Artificial Intelligence Generated Content (AIGC), conditional multimodal synthesis technologies (e.g., text-to-image, text-to-video, text-to-audio, etc) are gradually reshaping the natural content in the real world. The key to multimodal synthesis technology is to establish the mapping relationship between different modalities. Brain signals, serving as potential reflections of how the brain interprets external information, exhibit a distinctive One-to-Many correspondence with various external modalities. This correspondence makes brain signals emerge as a promising guiding condition for multimodal content synthesis. Brian-conditional multimodal synthesis refers to decoding brain signals back to perceptual experience, which is crucial for developing practical brain-computer interface systems and unraveling complex mechanisms underlying how the brain perceives and comprehends external stimuli. This survey comprehensively examines the emerging field of \textit{\textbf{AIGC}-based \textbf{Brain}-conditional Multimodal Synthesis}, termed \textbf{AIGC-Brain}, to delineate the current landscape and future directions. To begin, related brain neuroimaging datasets, functional brain regions, and mainstream generative models are introduced as the foundation of AIGC-Brain decoding and analysis. Next, we provide a comprehensive taxonomy for AIGC-Brain decoding models and present task-specific representative work and detailed implementation strategies to facilitate comparison and in-depth analysis. Quality assessments are then introduced for both qualitative and quantitative evaluation. Finally, this survey explores insights gained, providing current challenges and outlining prospects of AIGC-Brain. Being the inaugural survey in this domain, this paper paves the way for the progress of AIGC-Brain research, offering a foundational overview to guide future work. A webpage associated with this survey is available at: \url{https://github.com/MichaelMaiii/AIGC-Brain}.
\end{abstract}


\section{Introduction}
\label{sec:intro}

\begin{figure}[!t]
\centering
\includegraphics[width=\linewidth]{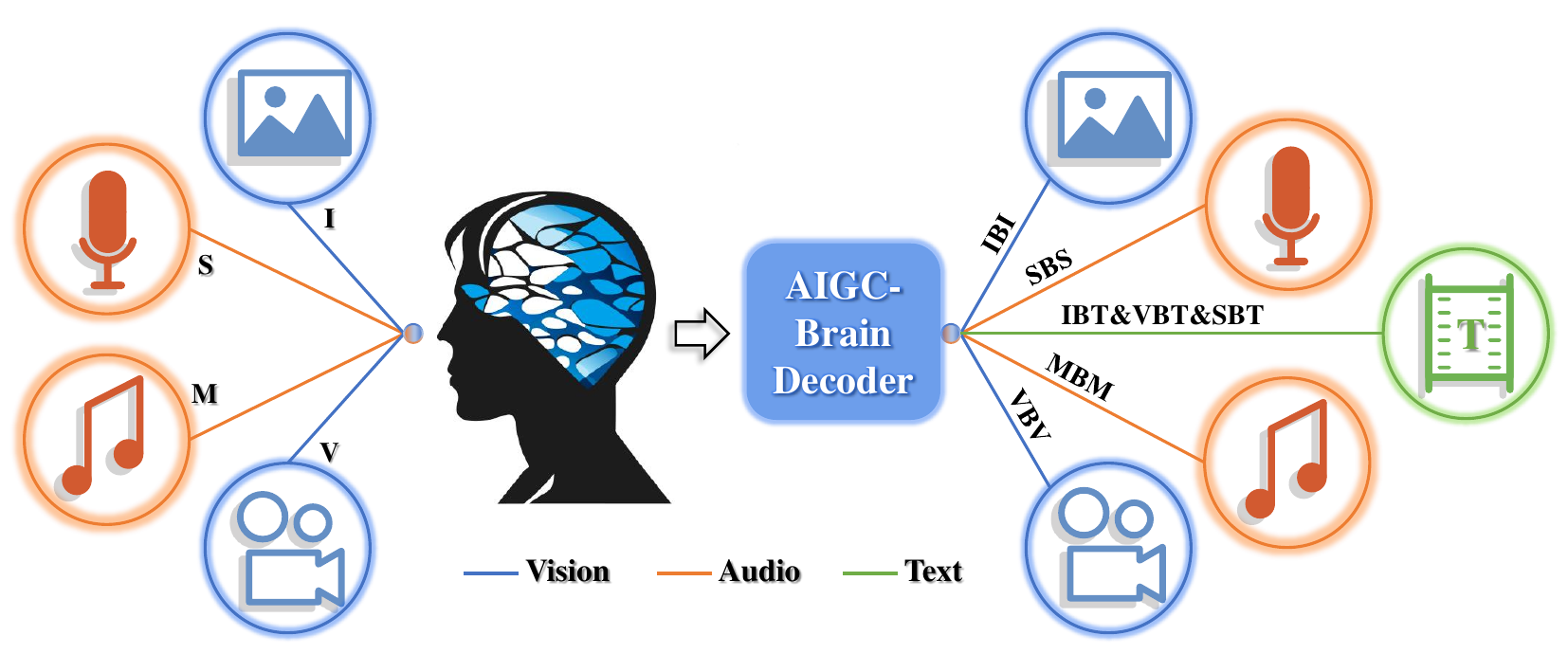}
\caption{Brain-Conditional Multimodal Synthesis via AIGC-Brain Decoder. Sensory stimuli comprising visual stimuli (Image (I), Video (V)) and audio stimuli (Music (M), Speech/Sound (S)) from the external world are first encoded to non-invasive brain signals (EEG, fMRI, or MEG) and then decoded back to perceptual experience via the AIGC-Brain decoder. This survey focuses on passive brain-conditional multimodal synthesis tasks including Image-Brain-Image (IBI), Video-Brain-Video (VBV), Sound-Brain-Sound (SBS), Music-Brain-Music (MBM), Image-Brain-Text (IBT), Video-Brain-Text (VBT), and Speech-Brain-Text (SBT), where IBI refers to image synthesis tasks conditioned on brain signals evoked by image stimuli.}
\label{fig_1}
\end{figure}

Understanding complex mechanisms by which the brain perceives the world is crucial for unraveling the mysteries of human perception\cite{audiovisual-perception, gilbert2007brain}. Sensory information is transformed into neural signals and transmitted through the nervous system to various brain regions. The correlation between external stimuli and brain signals provides valuable insights into the neural processes underlying perception\cite{fMRI-neurophysiological, EEG-perception, fMRI-visual-perception}. Perception refers to the process by which individuals receive, organize, and interpret sensory stimuli to make sense of the external world. It is not merely the passive reception of sensory inputs; rather, it involves complex cognitive processes that contribute to shaping subjective experience. Hence, decoding brain signals back to perception (e.g., corresponding vision, audio, and descriptive semantic text) holds significance in unraveling the complex mechanisms underlying perception. Moreover, brain perception decoding contributes to developing practical Brain-Computer Interface (BCI) systems. This technological advancement can potentially enhance communication between the brain and external devices, paving the way for novel applications in neuroprosthetics\cite{BCI-neuro-stroke, BCI-neuro}, virtual/augmented reality\cite{BCI-VR, BCI-AR}, and assistive\cite{BCI-Assist-Eating, BCI-Assist-Arm} technologies.

Neuroimaging technologies (e.g., functional Magnetic Resonance Imaging (fMRI), Electroencephalography (EEG), and Magnetoencephalography (MEG)) offer a window into the intricate neural activity associated with perception experiences\cite{EEG,fMRI,MEG}. Specifically, fMRI captures changes in blood flow with high spatial resolution, making it proficient in localizing brain activity but having limited temporal resolution. EEG records electrical activity on the scalp, providing excellent temporal resolution and sensitivity to neural processes but lacking spatial precision. MEG measures magnetic fields generated by neural activity, offering a balanced combination of spatial and temporal resolution. Neuroimaging data lays the foundation for brain decoding that facilitates the exploration of functionalities and interrelations of various brain regions, shedding light on the operational mechanisms by which the brain perceives and comprehends the external world\cite{Neuroimaging-Brain, neuroimaging-development}.

Generative models have evolved from deterministic Autoencoders (AEs)\cite{Autoencoder} to probabilistic models like Variational Autoencoders (VAEs)\cite{VAE}, Autoregressive Models, Generative Adversarial Networks (GANs)\cite{GANs}, and the current hottest Diffusion Models (DMs)\cite{DDPM, DM2, DM3}. Each development phase represents a significant advancement, incorporating probabilistic aspects, sequential dependencies, adversarial training, and iterative denoising processes, showcasing the continuous innovation in generative modeling. These methods have succeeded in various domains, including image, video, text, and audio synthesis. Conditional generative models extend the concept by introducing conditional information into the generation process, rather than just generating samples randomly from the data distribution\cite{CDM1, CDM2, CDM3}. Specifically, image-to-image translation\cite{I2I-CGAN} and style transfer\cite{ST-CGAN} within the image modality are based on conditional generative models. Moving further, conditional multimodal generative models integrate information from multiple modalities, including one-to-one synthesis (e.g., text-to-image\cite{SD}, text-to-video\cite{stable-video}, text-to-audio\cite{audioldm}, and image-to-text\cite{blip2}, audio-to-video\cite{aud2vid}), one-to-any synthesis (e.g., text-to-image\&video\cite{nuwa}), any-to-one synthesis (e.g., text\&image-to-video\cite{videopoet}), and a unified model for any-to-any synthesis (e.g., text\&image-to-video\&speech\cite{girdhar2023imagebind, Unified-IO}).

The key to multimodal synthesis technology is to establish the mapping relationship between different modalities. Brain signals, serving as potential reflections of how the brain interprets external information, exhibit a distinctive One-to-Many correspondence with various external modalities. This correspondence makes brain signals emerge as a promising guiding condition for multimodal content synthesis\cite{Mind-Reader, BrainSD, Mind-Vis}. 
This survey comprehensively examines the emerging field of AIGC-based brain-conditional multimodal synthesis, termed AIGC-Brain, to delineate the current landscape and future direction for further work.
Fig.\ref{fig_1} illustrates the pipeline of brain-conditional multimodal synthesis via the AIGC-Brain decoder. Sensory stimuli comprising visual stimuli (i.e., image (I) and video (V)) and audio stimuli (i.e., sound (S), speech (S), and music (M)) from the external world are first encoded to non-invasive brain signals (EEG, fMRI, or MEG) and then decoded back to perceptual experience via the AIGC-Brain decoder. Note that this survey focuses on passive brain decoding tasks where brain signals are evoked from sensory stimuli. So far, various AIGC-Brain decoding tasks have been explored, including Image-Brain-Image (IBI), Video-Brain-Video (VBV), Sound-Brain-Sound (SBS), Music-Brain-Music (MBM), Image-Brain-Text (IBT), Video-Brain-Text (VBT), and Speech-Brain-Text (SBT). Specifically, IBI refers to image synthesis tasks conditioned on brain signals evoked by image stimuli, VBV refers to video synthesis tasks conditioned on brain signals evoked by video stimuli, and so on. AIGC-Brain refers to decoding brain signals back to perception information (e.g., corresponding vision, audio, and semantic text) that contributes to developing practical BCI systems and unraveling how the brain process and understand the external world.

This survey delves into various aspects of AIGC-Brain, concentrating on \textit{passive synthesis tasks} from non-invasive brain signals evoked by sensory stimuli while other AIGC-Brain tasks such as \textit{active synthesis tasks} and \textit{invasive synthesis tasks} are introduced briefly. Note that this survey concentrates on generative tasks, excluding other research areas like retrieval\cite{EEG-retrieval}, modality matching\cite{matching-speech}, and classification\cite{BraVL, palazzo2020decoding}, as they are beyond the scope of this paper. This manuscript contributes significantly to the field in the following aspects: 
\begin{itemize}
    \item \textbf{Foundations:} We meticulously summarize related brain neuroimaging datasets, functional brain regions, and mainstream generative models as the foundation of AIGC-Brain decoding and analysis.
    \item \textbf{Methodology Taxonomy:} We systematically categorize AIGC-Brain decoding models based on shared characteristics in their implementation architectures, emphasizing workflows, intrinsic mapping relations, strengths, and weaknesses of each type of method. This methodology taxonomy serves to delineate the current methodological landscape of AIGC-Brain.
    \item  \textbf{Task-specific Implementation:} This survey presents detailed implementation strategies for various AIGC-Brain tasks, showcasing representative work and pipelines, to facilitate comparison and in-depth analysis of technological trends, task-specific characteristics, and preferences.
    \item \textbf{Quality Assessment:} This survey provides an overview of task-specific quality assessments in AIGC-Brain to facilitate the evaluation of synthesis results qualitatively and quantitatively.
    \item \textbf{Comprehensive Insights: }This work concludes with a concise summary of current challenges in AIGC-Brain and provides insightful perspectives for future research directions. 
\end{itemize}

The remainder of this survey is organized as follows. Section \ref{sec:dataset} and Section \ref{sec:region} provides a comprehensive overview and description of brain neuroimaging datasets and related functional brain regions in AIGC-Brain, respectively. Section \ref{sec:decoder} introduces mainstream generative models as the methodology foundation of AIGC-Brain. Section \ref{sec:method} provides a comprehensive methodology taxonomy for AIGC-Brain decoding models and Section \ref{sec: tasks} presents task-specific representative work and detailed implementation strategies. Section \ref{sec: results} reviews task-specific quality assessment with both qualitative and quantitative metrics. In Section \ref{sec: prospects}, we conclude this survey and discuss the main challenges and prospects for AIGC-Brain.

\begin{table*}[htbp]
  \centering
  \caption{Neuroimaging Datasets for AIGC-Brain Decoding Tasks.}
    \renewcommand{\arraystretch}{1.3}
    \resizebox{\textwidth}{!}{
    \begin{threeparttable}
    \begin{tabular}{c|c|c|m{16.78em}|m{8.665em}|c|m{10em}}
    \thickhline
     Modality     & Dataset & Tpye  & Experimental Paradigm & ROI / Channel &Sub & AIGC-Brain Task \\
    \hline
    \multirow{15.5}{*}{Image} & NSD\cite{NSD}   & fMRI  & Viewing 73,000 (72,000/1,000) natural scene images from COCO. & V1, V2, V3, V4, LOC, FFA, PPA & 8     & IBI \cite{BrainSD, BrainSD-TGD, UniBrain, BrainDiffuser, MindDiffuser, MindEye, BrainClip, BrainSCN, BrainCaptioning}  \newline IBT \cite{UniBrain, BrainCaptioning, DreamCatcher}  \\
\cline{2-7}          & GOD\cite{GOD}   & fMRI  & Viewing 1,250 (1,200/50)  natural images from 200-class of ImageNet. & V1, V2, V3, V4, LOC, FFA, PPA & 5     & IBI \cite{BrainClip, Mind-Vis, VQ-fMRI, BrainHVAE, BrainICG, BrainHSG, SBD, LEA, CMVDM, CAD, SSNIR, SSNIR-SC, BrainSSG, BrainBBG, BrainDVG, BrainDCG}  \newline IBT \cite{GIC-RL} \\
\cline{2-7}          & BOLD\cite{BOLD} & fMRI  & Viewing 4,916 distinct scene images from SUN, COCO, and ImageNet. & PPA, RSC, OPA, EVC & 4     & IBI \cite{Mind-Vis, LEA, CMVDM, CAD}  \\
\cline{2-7}          & DIR\cite{DIR}   & fMRI  & Viewing (or imaging) natural images (Same as GOD), artificial shapes (0/40), and alphabetical letters (0/10). & V1, V2, V3, V4, LOC, FFA, PPA & 3     & IBI \cite{DIR, E-DIR} \\
\cline{2-7}          & Vim-1\cite{Vim-1} & fMRI  & Viewing 1,870 (1,750/120) grayscale natural images. & V1, V2, V3, V4, LO & 2     & IBI \cite{SSNIR, SSNIR-SC, BrainDCG, BrainBG} \\
\cline{2-7}          & Faces\cite{Faces}  & fMRI  & Viewing 108 (88/20) face images from CelebA. & N/A   & 4     & IBI \cite{Faces} \\
\cline{2-7}          & OCD\cite{OCD}   & fMRI  & Viewing 2,750 (2,250/250/250) natural images from 5-class of ImageNet. & V1, V2, V3, LVC, HVC, VC & 5     & IBT \cite{GIC-PTL, GIC-CT} \\
\cline{2-7}          & EEG-VOA\cite{EEG-VOA} & EEG   & Viewing 2,000 (1,600/200/200) natural images from 40-class of ImageNet. & 128 EEG Channels & 6     & IBI \cite{DreamDiffusion, NeuroImagen, DM-RE2I, Brain2Image, NeuroVision, EEG-VGD, EEG-GAN} \\
\cline{2-7}          & MEG-Things\cite{Things-MEG} & MEG   &  Viewing 22,448 object images from 1,854-class from the THINGS database & 272 MEG Channels & 4    & IBI \cite{MEG-BD} \\
    \hline
    \multirow{3.5}{*}{Video} & DNV\cite{DNV} & fMRI  & Viewing 972 (374/598) natural color video clips from VideoBlocks and Youtube. & V1, V2, V3, V4, LO, MT, FFA, PPA, LIP, TPJ, PEF, FEF & 3     & VBV \cite{Mind-Video, SSRNM, f-CVGAN, BrainViVAE, DNV} \\
\cline{2-7}          & VER\cite{VER} & fMRI  & Viewing 12,600 (7,200/5,400) color natural clips from the Apple Quick-Time HD gallery and YouTube. & V1, V2, V3, V3A, V3B & 3     & IBT \cite{DSR, GNLD} \\
    \hline
    \multirow{3.5}{*}{\tabincell{c}{Video\\ \& \\Speech}} & STNS\cite{STNS} & fMRI  & Viewing and listening videos from 30 episodes of BBC’s Doctor Who. & V1, V2, V3, MT, AC, FFA, LOC, OFA & 1     & VBV \cite{Brain2Pix} \\
\cline{2-7}
    & CLSR\cite{CLSR} & fMRI  & Viewing silent video clips from animated films. \& Listening 82 stories from The Moth Radio Hour and Modern Love.  & AC, Broca, sPMv & 3     & VBT \& SBT \cite{CLSR} \\
    \hline
    \multirow{4.5}{*}{\tabincell{c}{Sound\\ \& \\Speech}} & BSR\cite{BSR} & fMRI  & Listening 1,250 (1,200/50) sound clips from VGGSound test dataset. & A1, LBelt, Pbelt, A4, A5 & 5     & SBS \cite{BSR} \\
\cline{2-7}          & Narratives\cite{Narratives} & fMRI  & Listening 27 diverse naturalistic spoken stories. & A1, Mbelt, LBelt, Pbelt, RI & 345   & SBT \cite{UniCoRN} \\
\cline{2-7}          & ETCAS\cite{ETCAS} & EEG   & Listening continuous speech audio from the TIMIT dataset. & 24 EEG Channels & 50 & SBS \cite{ETCAS} \\
    \hline
    \multirow{3}{*}{Music} & MusicGenre\cite{MusicGenre} & fMRI  & Listening 540 (480/60) music pieces from 10 music genres. & N/A   & 5     & MBM \cite{Brain2Music} \\
\cline{2-7}          & MusicAffect\cite{MusicAffect} & fMRI\&EEG & Listening Synthetic music and Classical music clips. & N/A \& \newline{} 31 EEG Channels & 21    & MBM \cite{NDMusic} \\
    \thickhline
    \end{tabular}%
    \begin{tablenotes} 
	\item[1] Public Large Dataset: COCO\cite{COCO}, ImageNet\cite{ImageNet}, SUN\cite{SUN}, CelebA\cite{CelebA}, Things\cite{Things}, VGGSound\cite{VGGSound}, TIMIT-\url{https://catalog.ldc.upenn.edu/docs/LDC93S1/TIMIT.html}
        \item[2] Website: VideoBlocks-\url{https://www.videoblocks.com}, Youtu-\url{https://www.youtube.com},  Apple Quick-Time HD Gallery-\url{http://trailers.apple.com}
    \end{tablenotes} 
    \end{threeparttable}
    }
  \label{tab-datasets}%
\end{table*}%

\section{Neuroimaging Datasets}
\label{sec:dataset}
Neuroimaging data (e.g., fMRI, EEG, MEG) lays the foundation for brain decoding research that facilitates the exploration of functionalities and interrelations of various brain regions, shedding light on the operational mechanisms by which the brain perceives and comprehends the external world\cite{Neuroimaging-Brain, neuroimaging-development}.
Constructing neuroimaging datasets involves costly equipment, technical challenges, as well as ethical and privacy concerns. We greatly appreciate the authors' contributions to existing open-source datasets, as they offer a valuable resource for neuroscience and medical research, thereby enhancing our comprehension of the human brain. In this section, we summarize public neuroimaging datasets in terms of brain responses to visual and auditory stimuli that have been used for AIGC-Brain decoding tasks. As shown in Table \ref{tab-datasets}, neuroimaging datasets are divided according to their stimulus modality. For each dataset, the reference source (Dataset), data type (Type), specific experimental paradigm,  fMRI region-of-interest or EEG/MEG channels (ROI/Channel), number of subjects (Sub), and associated AIGC-Brain tasks are presented.

\subsection{Image Datasets}
\label{subsec: image-dataset}
While there exist a variety of neuroimaging datasets used for brain-conditional image reconstruction, such as binary contrast patterns (BCP)  \cite{BCP}, 6-9 dataset of handwritten digits \cite{6-9}, BRAINS dataset of handwritten characters \cite{BRAINS}, we focus on the datasets with a higher level of perceptual complexity of presented stimuli: datasets of faces, grayscale natural images, and natural images. 

The Natural Scenes Dataset (\textbf{NSD}\footnote{https://naturalscenesdataset.org})\cite{NSD} is currently the largest fMRI-Image dataset gathered from 8 subjects viewing 73,000 natural scene images from the COCO dataset \cite{COCO}. Researchers usually focus on four subjects (Sub-1, Sub-2, Sub-5, and Sub-7) who finished all viewing trials. The test images (1,000) remain consistent across all subjects, whereas distinct training images (9,000x8) are adopted. NSD is currently one of the most popular image datasets and has been used for IBI\cite{BrainSD, BrainSD-TGD, UniBrain, BrainDiffuser, MindDiffuser, MindEye, BrainClip, BrainSCN, Mind-Reader, BrainCaptioning, Dream} and IBT\cite{UniBrain, BrainCaptioning, DreamCatcher} tasks. The corresponding captions in the COCO dataset can serve as semantic descriptions of images. The Generic Object Decoding (\textbf{GOD}\footnote{https://github.com/KamitaniLab/GenericObjectDecoding}) dataset \cite{GOD} is an fMRI dataset gathered from 5 subjects viewing 1,250 natural object images from 200 classes of ImageNet. The training set consists of 1,200 images from 150 classes, while the test set comprises only 50 images from 50 classes. Therefore, brain decoding on GOD involves zero-shot learning of category information, and it may fall short in complex scenes with multiple objects. Furthermore, current GOD-based research generally uses BLIP\cite{BLIP} to generate semantic captions to obtain semantic information about images. GOD is also one of the most popular image datasets and has been used for IBI \cite{BrainClip, Mind-Vis, VQ-fMRI, BrainHVAE, BrainICG, BrainHSG, SBD, LEA, CMVDM, CAD, SSNIR, SSNIR-SC, BrainSSG, BrainBBG, BrainDVG, BrainDCG, DBDM} and IBT \cite{GIC-RL} tasks. 
The Deep Image Reconstruction (\textbf{DIR}) dataset \cite{DIR} has the same natural image dataset as GOD, with 40 artificial shapes, and 10 alphabetical letters added during the testing sessions.
The Brain, Object, Landscape Dataset (\textbf{BOLD}) \cite{BOLD} and the grayscale natural \textbf{Vim-1} dataset \cite{Vim-1} are generally accompanied by the GOD dataset to verify the generalization ability of the AIGC-Brain decoder. Similar to GOD, the Object Category Decoding (\textbf{OCD}) dataset\cite{OCD} is gathered from 5 subjects viewing 2,750 (train/valid/test: 2,250/250/250) natural object images from 5 classes of ImageNet. However, it is currently only used for IBT tasks\cite{GIC-PTL, GIC-CT}. The \textbf{Faces} dataset \cite{Faces} is currently the only fMRI dataset focused on face image reconstruction.
The EEG Visual Object Analysis dataset (\textbf{EEG-VOA}\footnote{https://github.com/perceivelab/eeg\_visual\_classification})\cite{EEG-VOA} is collected using EEG equipment with 128 channels placed following the international 10–10 system. Similar to GOD, image stimuli in EEG-VOA are from 40 classes of ImageNet, but it only uses a random 8 (train):1 (valid):1 (test) ratio to split the dataset. EEG-VOA is the most popular EEG dataset that has been used for IBI tasks\cite{DreamDiffusion, NeuroImagen, DM-RE2I, Brain2Image, NeuroVision, EEG-VGD, EEG-GAN}. 
The \textbf{MEG-Things\footnote{https://things-initiative.org/}}\cite{Things-MEG} dataset is collected using MEG equipment with 272 MEG channels, while 4 subjects viewed 22,448 unique object images from 1,854 classes in the THINGS\cite{Things} database.

\subsection{Video Datasets}
\label{subsec: video-dataset}
The Dynamic Natural Vision (\textbf{DNV}) dataset \cite{DNV} is collected from 3 subjects viewing natural color video clips. DNV is currently the most popular fMRI-Video dataset for VBV tasks \cite{Mind-Video, SSRNM, f-CVGAN, BrainViVAE, DNV}. The Visual Experience Reconstruction (\textbf{VER}) dataset \cite{VER} only involves ROIs of the early visual cortex and it has only been used for the IBT task in recent years \cite{DSR, GNLD}.

\subsection{Video\&Speech Datasets}
\label{subsec: video-speech-dataset}
The Space-Time Natural Stimuli (\textbf{STNS}) dataset \cite{STNS} is gathered from single participant viewing and listening videos from episodes, involving ROIs of visual and auditory cortexes. STNS has been used for VBV tasks in \cite{Brain2Pix}. The Continous Language Semantic Reconstruction (\textbf{CLSR}) dataset provides fMRI-Speech data for training, and fMRI-Video data for cross-modal testing, involving ROIs of auditory and language cortexes. Data collection and semantic decoding tasks, i.e., SBT and VBT, are completed in \cite{CLSR}.

\subsection{Sound\&Speech Datasets}
\label{subsec: speech-dataset}
Narratives \cite{Narratives} is a huge fMRI-Speech dataset gathered from 345 subjects listening to diverse naturalistic spoken stories. It focuses on the early and higher auditory cortex and has been used for the SBT task\cite{UniCoRN}. The Brain Sound Reconstruction dataset (\textbf{BSR})\cite{BSR} is the fMRI-Sound dataset for the SBS task involving the early auditory cortex. The \textbf{ETCAS}\cite{ETCAS} dataset is collected from 50 subjects listening to continuous speech using EEG. Data collection and speech/sound reconstruction tasks are completed in their respective work\cite{BSR, ETCAS}.

\subsection{Music Datasets}
\label{subsec: music-dataset}
Reconstructing musical stimuli from brain signals is a more complex task that still needs more exploration. We only summarize two datasets that are used for the MBM task\cite{Brain2Music, NDMusic}. Specifically, the \textbf{MusicGenre}\cite{MusicGenre} dataset is gathered from 5 subjects listening to music pieces from 10 music genres. The \textbf{MusicAffect}\cite{MusicAffect} dataset is recorded via a joint EEG-fMRI imaging modality while subjects listen to music with different affective states.

\begin{figure}[!t]
\centering
\includegraphics[width=\linewidth]{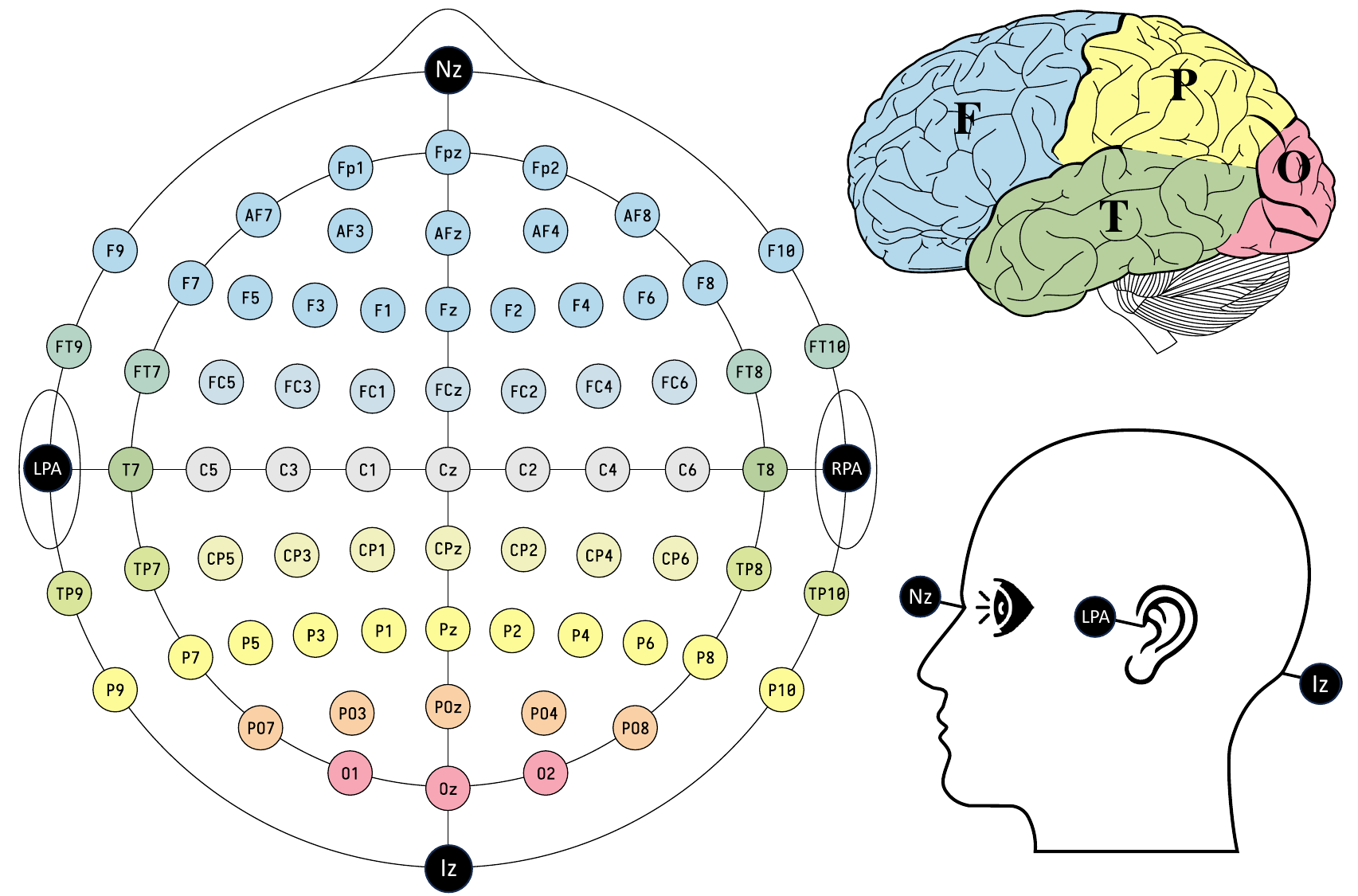}
\caption{EEG 10-10 channel system and brain regions. Pink: Occipital lobe; Green: Temporal lobe; Yellow: Parietal lobe; Blue: Frontal lobe.}
\label{fig_region}
\end{figure}

\section{Brain Regions}
\label{sec:region}
This section will describe the relevant brain regions responsible for visual, auditory, and language perception, as well as their locations in the brain, including the occipital (O), temporal (T), parietal (P), and frontal (F) lobes, as shown in Fig.\ref{fig_region}. EEG Electrodes, which are located in brain regions with the same color in Fig.\ref{fig_region}, are placed following the international 10-10 channel system. Moreover, the placement of MEG channels on the scalp is conceptually similar to that of EEG channels adhering to international channel systems.

\subsection{Visual Cortex}
\label{subsec: roi-vision}
The visual cortex is a critical part of the brain responsible for processing visual information from the eyes. It plays a fundamental role in visual perception, allowing us to see and understand the world around us.

\subsubsection{Early Visual Cortex}
The early visual cortex (EVC) is primarily responsible for processing basic visual information, including the detection of simple features like edges, colors, shapes, and motion\cite{tong2003primary}. Specifically, the primary visual cortex V1 and the secondary visual cortex V2 process basic visual information for recognizing simple geometric patterns, such as edges and orientations. V3 specializes in motion perception and the processing of motion-related visual stimuli. V4 is primarily associated with color and form processing, contributing to color perception and object recognition. The Middle Temporal area (MT), also known as V5, is crucial for motion processing, particularly the detection of object motion and motion direction. The Occipital Face Area (OFA) is specialized for processing facial features and is involved in the early stages of face perception.  

\subsubsection{Higher Visual Cortex}
Higher visual cortex (HVC) regions are involved in more advanced visual processing tasks, such as object recognition, facial recognition, scene perception, and the integration of complex visual information\cite{grill2004human}. Specifically, the Lateral Occipital Complex (LOC) is associated with the perception of complex visual shapes and structures for object recognition. The Fusiform Face Area (FFA) is specialized for recognizing and processing faces. The Parahippocampal Place Area (PPA) is responsible for recognizing scenes and spatial layouts. Similar to the PPA, the Occipital Place Area (OPA) processes scenes and places. The Retrosplenial Cortex (RSC) is involved in spatial navigation, memory, and scene processing, contributing to the creation of a mental map of the environment. The Lateral Intraparietal Area (LIP) is associated with the dorsal visual stream, focusing on spatial perception, attention, and eye movement control. The Temporoparietal Junction (TPJ) is involved in various functions, including social cognition, perspective-taking, and attention.

In summary, EVC regions (V1, V2, V3, V4, MT, OFA) primarily handle basic visual processing, while HVC regions (LOC, FFA, PPA, OPA, RSC, LIP, TPJ) are responsible for more advanced visual processing tasks (i.e., face recognition, scene perception, and social cognition). EVC is primarily located in the occipital (O) lobe of the brain. As information progresses from EVC to the HVC, the processing of visual information becomes more complex and abstract. As a result, HVC refers to regions situated at higher levels extending into the temporal (T) and parietal (P) lobes (Fig.\ref{fig_region}).

\subsection{Auditory Cortex}
\label{subsec: roi-audio}
The auditory cortex (AC) is a specific brain region responsible for processing auditory information, including the perception of sound, speech, music, and other auditory stimuli. It is typically located in the temporal (T) lobe (Fig.\ref{fig_region}). The early auditory cortex (EAC)\cite{primary-EAC}, including A1 (Primary Auditory Cortex), A4, and A5, focuses on fundamental auditory processing, such as pitch and frequency. In contrast, the higher auditory cortex (HAC)\cite{brewer2016maps}, comprising the Lateral Belt (LBelt), Posterior Belt (PBelt), Medial Belt (MBelt), and Rostral Intermediate (RI), is involved in more advanced functions like sound localization, integration of complex auditory stimuli, and the comprehensive processing of auditory information in conjunction with other sensory modalities. Together, these distinct areas work in concert to form a sophisticated system for perceiving and interpreting the diverse aspects of auditory stimuli.

\subsection{Language Cortex}
\label{subsec: roi-language}
The language cortex, which is typically located in the Frontal (F) lobe, refers to brain regions that are involved in language processing. It includes various regions responsible for both understanding and producing spoken and written language. Language processing is a complex and distributed function, and it engages several brain areas working in concert. The Broca's Area (Broca) plays a critical role in language production and speech\cite{musso2003broca}. It is associated with the ability to form grammatically correct sentences, coordinate the muscles necessary for speech (e.g., lips, tongue, and vocal cords), and generate fluent, coherent language. The Superior Ventral Premotor Speech Area (sPMv) is primarily involved in the motor planning and control of speech production\cite{ventral}. sPMv is crucial for coordinating the movements of the articulatory muscles, allowing us to produce the precise sounds and phonemes required for spoken language.

\section{Generative Models}
\label{sec:decoder}
AIGC-Brain aims to synthesize multimodal content conditioned on brain signals, as shown in Fig.\ref{fig_1}.  
The AIGC-Brain decoder is structured with two key elements: the modality matching network and the AIGC decoder. The modality matching network is commonly constructed using linear regressions or simple neural networks. The AIGC decoder typically employs deep generative models including Diffusion Models (DMs), Generative Adversarial Networks (GANs), Variational Autoencoders (VAEs), Convolutional Autoencoders (CAEs), and Autoregressive Models. This section will briefly introduce generative models that have been utilized for AIGC-Brain decoding.

\subsection{Diffusion Models}
\label{subsec: ldm}

Diffusion Models (DMs)\cite{DDPM} are a class of probabilistic generative models that leverage iterative denoising to transform a sampled variable from Gaussian noise into a sample adhering to the learned data distribution. The forward diffusion process gradually introduces Gaussian noise to the initial input $x_0$ at each time point, defined as $x_t = \sqrt{\alpha_t}x_0 + \sqrt{1-\alpha_t}\epsilon_t$. A neural network, termed the denoising U-Net\cite{U-Net}, is trained to perform the reverse diffusion process, predicting and removing noise from the noisy input to recover the original variables. The loss function for this process is expressed as:

\[
L_{DM}=E_{x_0, \epsilon\sim\mathcal N(0,1) ,t}\left [||\epsilon-\epsilon_\theta(x_t,t)||^2\right]
\]
where $\alpha$ controls the noise addition, $\epsilon$ represents true Gaussian noise, $\epsilon_\theta(\cdot)$ is the neural network predicting the noise, and $t\in\{1,2,...,T\}$ denotes the time step. Specifically, Guided Diffusion (GD) \cite{GD} is an instance of a diffusion model that incorporates classifier guidance for image synthesis.

Pixel space DMs come with significant computational costs. Latent Diffusion Models (LDMs), also termed Stable Diffusion (SD) \cite{SD}, overcome this challenge by compressing the input with an autoencoder $E(\cdot)$ trained on a large-scale image dataset to learn a compressed latent representation $z_0$ from image $x_0$ ($z_0 = E(x_0)$). The forward diffusion process in LDM is denoted as $z_t = \sqrt{\alpha_t}z_0 + \sqrt{1-\alpha_t}\epsilon_t$. The reverse diffusion process introduces additional conditions in a latent space, with an objective function defined as:

\[
L_{LDM}=E_{z_0, c, \epsilon\sim\mathcal N(0,1) ,t}\left [||\epsilon-\epsilon(z_t,t,\tau_\theta(c)||^2\right]
\]
where $\tau_\theta(c)$ is the conditioning input for U-Net. The innovative aspect of LDM lies in its ability to guide the inverse diffusion process using various conditions (e.g., labels, captions, images, and semantic maps). Conditioning is achieved by integrating conditions $\tau_\theta(c)$ within the cross-attention block of the denoising U-Net model. The denoised latent variable resulting from the reverse diffusion is then passed through the pretrained decoder $D(\cdot)$ to generate a high-quality image. 

ControlNet \cite{ControlNet} serves as an extra control neural network working with large pretrained text-to-image diffusion models (i.e., Stable Diffusion). Besides text conditions, additional spatial conditions (e.g., human pose, canny edge, and depth map) can be incorporated into diffusion models to further control image generation. Moreover, Versatile Diffusion (VD) \cite{VD} stands out as a multi-flow multimodal latent diffusion model with the capacity to generate diverse outputs, such as images and text. The guidance for this versatility comes from CLIP\cite{Clip} features derived from images, text, or combinations of both. VD serves as an advanced and adaptable model for generating diverse outputs in the context of image and text synthesis.


\subsection{Generative Adversarial Networks}
\label{subsec: GAN}
Generative Adversarial Networks (GANs)\cite{GANs} constitute a revolutionary approach to generative modeling. The central idea involves training a generator and discriminator simultaneously through adversarial training. The generator aims to produce synthetic data that is indistinguishable from real data, while the discriminator works to differentiate between real and generated samples. This adversarial game is formulated with the following objective function:

\[ \begin{aligned}
\min_G \max_D V(D, G) &= \mathbb{E}_{x \sim p_{\text{data}}(x)}[\log D(x)] \\
&   + \mathbb{E}_{z \sim p_z(z)}[\log(1 - D(G(z)))]
\end{aligned} \]

In the realm of image generation, GANs have achieved unparalleled success, generating high-quality, diverse images with realistic details. Their applications extend to various domains, including video synthesis (e.g., Vid2Vid\cite{vid2vid}, MoCoGAN\cite{MoCoGAN}, and TGAN\cite{TGAN}), audio generation (e.g., WaveGAN\cite{WaveGAN} and GAN-TTS\cite{GAN-TTS}), and image synthesis (e.g., DCGAN\cite{DCGAN}, StyleGAN\cite{StyleGAN}, StyleGAN2\cite{StyleGAN2}, BigGAN\cite{BrainBG}, PGGAN\cite{PGGAN}). Conditional GANs (cGANs) such as InfoGAN\cite{InfoGan}, ICGAN\cite{ICGAN}, CycleGAN\cite{CycleGAN}, and Pix2Pix\cite{pix2pix} are an extension of the traditional GANs that incorporate additional conditional information during the training and generation processes. GANs' ability to capture intricate patterns and produce compelling, lifelike content positions them at the forefront of generative modeling. While GANs offer remarkable advantages, such as the production of realistic samples and their versatility across domains, challenges persist, including training instability, convergence issues, and mode collapse.

\subsection{Variational Autoencoders}
\label{subsec: VAE}
Variational Autoencoders (VAEs)\cite{VAE} represents a prominent paradigm in generative modeling, combining variational inference and autoencoder architecture. The fundamental principle involves encoding input data $x$ into a probability distribution in the latent space, facilitating stochastic sampling and decoding for accurate reconstruction. The VAE objective function is defined as:
\[ \mathcal{L}_{\text{VAE}} = \mathbb{E}_{q(z|x)}[\log p(x|z)] - \text{KL}(q(z|x)||p(z)) \]
where the first term represents the reconstruction loss, and the second term is the Kullback-Leibler divergence between the distribution of latent variables given the data (\(q(z|x)\)) and a prior distribution (\(p(z)\)). 

As variants of VAEs, VQ-VAE\cite{VQVAE} introduces vector quantization to enhance discrete latent variable representations and is further improved by VQ-VAE2\cite{VQVAE2}. Conditional VAE (CVAE)\cite{CVAE} incorporates conditional information during the generation process, offering contextual considerations. VAEs excel in learning interpretable latent representations but may generate blurry samples. To address this, VAE-GAN\cite{VAE-GAN} combines VAEs with GANs to enhance sample quality by integrating adversarial training. This hybrid model holds promise in producing refined and diverse samples, overcoming VAE limitations.

\begin{figure*}[!t]
\centering
\includegraphics[width=\linewidth]{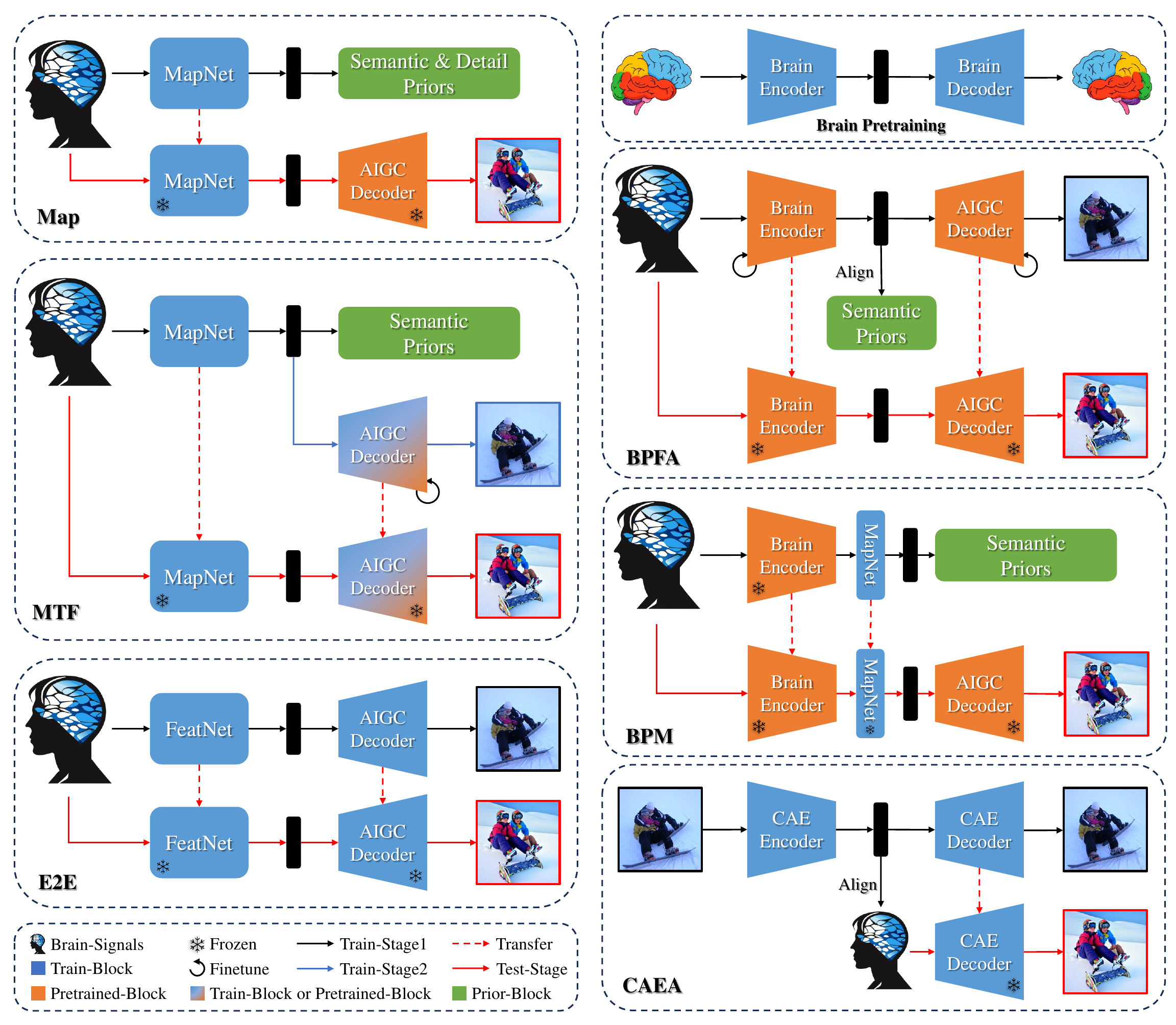}
\caption{Different types of methods for AIGC-Brain tasks. Map: Mapping; MTF: Map\&Train\&Finetune; E2E: End-to-End; BPM: Brain-Pretrain\&Map; BPFA: Brain-Pretrain\&Finetune\&Align; AEA: Auto-Encoder\&Align. Brain pretraining on large-scale neuroimaging datasets (e.g., HCP\cite{HCP} and MOABB\cite{MOABB}) is the first stage of BPFA and BPM methods.}
\label{fig_method}
\end{figure*}

\begin{table*}[htbp]
  \centering
  \caption{Different Method Types and Related Information for AIGC-Brain Tasks.}
  \renewcommand{\arraystretch}{1.3}
  \resizebox{\textwidth}{!}{
    \begin{tabular}{c|l|m{11.89em}|l|l|m{10.335em}}
    \thickhline
    Method Type & Mapping Relation & Description & Strength & Weakness & References \\
    \hline
    Map & Brain-Prior & Map brain signals to semantic or detail priors of the pretrained AIGC decoder. & \tabincell{l}{Easy training \\ Flexible implementation \\ High fidelity} & Bias superposition & IBI:\cite{BrainSD, BrainSD-TGD, UniBrain, BrainDiffuser, MindDiffuser, MindEye, BrainClip, BrainCaptioning, SBD, NeuroImagen, BrainSCN, Mind-Reader, BrainICG, BrainHSG, BrainSSG, BrainBBG, BrainDCG, DIR, BrainBG, Faces, VQ-fMRI, BrainHVAE, DBDM, Dream, MEG-BD} VBV: \cite{BrainViVAE} SBS:\cite{BSR} MBM:\cite{Brain2Music} IBT:\cite{UniBrain, DreamCatcher, GIC-RL, DSR, GNLD} VBT\&SBT:\cite{CLSR}\\
    \hline
    BPM & Brain-Prior & Pretrain on brain signals and then map to semantic priors of the pretrained AIGC decoder. & High semantic fidelity & \tabincell{l}{Bias superposition \\ Hard pretraining} & IBI:\cite{LEA} SBT:\cite{UniCoRN} \\
    \hline
    BPFA & Brain-Prior-Stimuli & Pretrain on brain signals and then finetune the pretrained AIGC decoder with/without semantic prior alignment. & High semantic fidelity & \tabincell{l}{Hard pretraining\\Hard finetuning} & IBI:\cite{Mind-Vis, CMVDM, CAD, DreamDiffusion} VBV:\cite{Mind-Video}\\
    \hline
    MTF   & Brain-Prior-Stimuli & Map brain signals to semantic priors, then train from scratch or finetune the AIGC decoder.  & High semantic fidelity & \tabincell{l}{Hard training \\ Hard finetuning} &  IBI:\cite{DM-RE2I, NeuroVision, EEG-GAN, EEG-VGD, Brain2Image, BrainDVG} IBT:\cite{GIC-PTL}\\
    \hline
    E2E & Brain-Stimuli & Map brain signals directly to sensory stimuli. & \tabincell{l}{Simplicity \\ Real-time application} & \tabincell{l}{Poor interpretability \\ Low semantic} & IBI:\cite{E-DIR} VBV:\cite{f-CVGAN, Brain2Pix} IBT:\cite{GIC-CT} MBM:\cite{NDMusic}  \\
    \hline
    CAEA & Brain-Stimuli & Map brain signals directly to sensory stimuli in a deterministic manner. & \tabincell{l}{High detail fidelity} & \tabincell{l}{Low realistic \\ Low semantic} & IBI:\cite{SSNIR, SSNIR-SC, SSDR} VBV:\cite{SSRNM, DNV} \\
    \thickhline
    \end{tabular}%
  \label{tab-summary}%
  }
\end{table*}%

\subsection{Convolutional Autoencoders}
\label{subsec: CAE}
The Convolutional Autoencoder (CAE)\cite{CAE}, operating as a deterministic generative model, leverages convolutional neural networks (CNNs) for both encoding and decoding processes. The encoder utilizes convolutional layers to extract hierarchical features from the input data, transforming it into a latent representation. The decoder, which typically employs deconvolutional layers or transposed convolutions, then reconstructs the input from this latent space. This deterministic architecture finds extensive application in the generation domain, excelling in tasks such as image denoising\cite{CAE-Denoising}, image compression\cite{CAE-Compression}, and image-to-image translation\cite{CAE-I2I}.

One notable distinction from probabilistic generative models lies in the deterministic nature of CAE. Unlike probabilistic models that provide a distribution of possible outputs, the deterministic model ensures a single, consistent reconstruction for a given input. This characteristic makes it computationally efficient and suitable for applications where deterministic outputs are preferred. While CAE exhibits advantages in capturing complex patterns and spatial dependencies, it may face challenges such as overfitting training data, limited semantic expressiveness, and poor diversity compared to its probabilistic counterparts. 

\subsection{Autoregressive Models}
\label{subsec: Autoregressive}
Autoregressive models have evolved substantially, progressing from Gate Recurrent Unit (GRU)\cite{GRU} and Long-Short Term Memory (LSTM)\cite{LSTM} architectures to the transformative Transformer\cite{Transformer} model. Originally designed to address sequential dependencies, GRU and LSTM introduced gating mechanisms in recurrent neural networks (RNNs). The Transformer architecture revolutionized sequence modeling by leveraging self-attention mechanisms, allowing for parallelization and capturing long-range dependencies efficiently. 

Across diverse domains, autoregressive models have demonstrated remarkable efficacy. In natural language processing, particularly text-related tasks, models like GPT (Generative Pre-trained Transformer) series (i.e., GPT\cite{gpt}, GPT-2\cite{gpt2}, GPT-3\cite{gpt3}, GPT-4\cite{gpt4}) showcase state-of-the-art language understanding and generation capabilities. Adapting to image generation, autoregressive structures, exemplified by PixelRNN\cite{PixelRNN}, PixelCNN\cite{PixelCNN}, and transformer-based models (e.g., Taming\cite{Taming}, iGPT\cite{iGPT}, CogView\cite{Cogview}, and ImageBart\cite{Imagebart}),  predict pixel values or tokens sequentially. In audio processing, autoregressive models such as WaveNet\cite{wavenet} excel in generating high-fidelity audio waveforms.

Autoregressive models excel at capturing sequential dependencies and possess the ability to generate diverse and realistic outputs, notably in text, image, and audio generation. However, challenges persist, with sequential processing impacting training and inference speed. Additionally, capturing extensive dependencies and mitigating vanishing or exploding gradients pose ongoing challenges.

\section{AIGC-Brain Methodology Taxonomy}
\label{sec:method}
Currently, the advancement in AIGC-Brain is constrained by challenges such as the inherent noise and variability in brain signals, the limited availability of brain-stimulus data pairs, the modality matching between brain and external modalities, and the integration of multimodal generative models. 
In this survey, we summarize the recent AIGC-Brain decoding models and categorize them into six types of methods according to their architecture: Map, Brain-Pretrain\&Map (BPM), Brain-Pretrain\&Finetune\&Align (BPFA), Map\&Train\&Finetune (MTF), End-to-End (E2E), and Convolutional-Autoencoder\&Align (CAEA). Table \ref{tab-summary} summarizes the mapping relation, descriptions, strengths, and weaknesses of each type of method, along with the corresponding AIGC-Brain tasks and models based on these methods. Furthermore, since all methods have been applied to the IBI task, we draw the framework diagram of each method for comparison by taking the IBI task as an example, as shown in Fig.\ref{fig_method}. Note that the pretrained AIGC decoders are pretrained on mainstream data such as text, images, videos, and audio. Within the brain-pretraining stage, the brain encoder and decoder are pretrained on large-scale neuroimaging datasets, i.e., fMRI-based dataset HCP\cite{HCP}, and EEG-based dataset MOABB\cite{MOABB} used in DreamDiffusion\cite{DreamDiffusion}. 

\subsection{Map}
\label{subsec: mapping}
The Map method refers to mapping brain signals to low-level detail priors or high-level semantic priors of the pretrained AIGC decoder (Fig.\ref{fig_method}-Mapping). The Map method only constructs the brain-prior connection between brain signals and prior knowledge, which leads to a problem that biases generated in the mapping space tend to stack into the generation space leading to semantic ambiguity (Table \ref{tab-summary}). However, in this AIGC era, the Map method is currently the most popular method that covers all AIGC-Brain tasks due to its characteristics of easy training, flexible implementation, and high generation quality. 
In the Map method, the purpose of the mapping network is to map brain signals into prior spaces of the pretrained AIGC decoder, that is, to establish the connection between brain signals and priors (i.e., detail and/or semantic priors). Upon this, specific modality content can be synthesized by the pretrained AIGC decoder.
Overall, the research cores of the Map method lie in two aspects as follows: i) how to choose and integrate powerful pretrained AIGC decoders to improve multimodal synthesis capabilities, and ii) how to design mapping networks to reproduce more accurate prior information from brain signals.

\subsection{Brain-Pretrain\&Finetune\&Align}
\label{subsec: BPFA}
The Brain-Pretrain\&Finetune\&Align (BPFA) method consists of two stages: i) Brain pretraining using large neuroimaging datasets to learn latent features from brain signals, and ii) Finetune the pretrained AIGC decoder using latent features with/without semantic prior alignment (Fig.\ref{fig_method}-BPFA). When finetuning, BPFA is equivalent to transferring the semantic prior knowledge of the pretrained AIGC decoder (e.g., class priors from the label-to-image decoder) to brain latent features under the objective constraints of sensory stimuli, thus constructing the brain-prior-stimuli connection. BPFA addresses bias superposition in the Map method, but the issue of semantic ambiguity still remains. The problem with BPFM is that both stage-1 pretraining and stage-2 finetuning are relatively difficult to achieve. Therefore, the use of BPFM mainly focuses on a small number of IBI and VBV tasks (Tabel \ref{tab-summary}). 

\subsection{Brain-Pretrain\&Map}
\label{subsec: BPM}
The Brain-Pretrain\&Map (BPM) method follows the idea of mapping but uses brain latent features instead of brain signals. BPM first performs brain pretraining to extract latent features, just as done in stage-1 of BPFA. Next, latent features are mapped to semantic priors of the pretrained AIGC decoder (Fig.\ref{fig_method}-BPM). It has not been proven that latent features are inherently superior to raw brain signals, as the performance of downstream tasks is influenced by the nature of pretraining tasks. Furthermore, pretraining on large neuroimaging datasets is a time-consuming and computationally expensive task. As a result, only a few studies and tasks have used the BPM method (Table \ref{tab-summary}). 

\subsection{Map\&Train\&Finetune}
\label{subsec: MTF}
The Map\&Train\&Finetune (MTF) method also establishes a connection between the brain, prior, and stimuli, but in a different way. In the first stage, MTF maps brain signals to semantic priors, as in the Map method. In the second stage, MTF prefers to train the AIGC decoder from scratch or finetune the pretrained AIGC decoder (Fig.\ref{fig_method}-MTF). However, training or finetuning deep generative models is challenging due to the scarcity of brain-stimuli pair data and the low signal-to-noise ratio of non-invasive brain signals (Tabel \ref{tab-summary}). We found that most typical models for EEG-based IBI tasks are dependent on this method, except the state-of-the-art methods like DreamDiffusion\cite{DreamDiffusion} and NeuroImagen\cite{NeuroImagen}. 

\subsection{End-to-End}
\label{subsec: E2E}
The End-to-End (E2E) method refers to mapping brain signals directly to sensory stimuli by training the AIGC decoder from scratch with/without the feature extractor. E2E establishes the brain-stimuli connection by mapping brain signals directly to sensory stimuli (Fig.\ref{fig_method}-E2E). It provides a simple architecture suitable for real-time applications. However, the scarcity of data pairs poses a challenge to effectively investigate the semantic information of brain signals, and the "black-box" design of its architecture hinders interpretability (Tabel \ref{tab-summary}).

\subsection{Convolutional-Autoencoder\&Align}
\label{subsec: CAEA}

The Convolutional-Autoencoder\&Align (CAEA) method establishes a direct and deterministic connection between brain signals and stimuli by aligning brain signals with the latent features of CAE (Fig.\ref{fig_method}-CAEA). Currently, CAEA has been applied to visual-related tasks (i.e., IBI and VBV), and modality alignment in CAEA mainly depends on supervised and self-supervised learning\cite{SSNIR, SSNIR-SC}. The CAE decoder usually consists of deconvolutional neural networks (DeCNNs) that prioritize generating pixels and outlines of visual stimuli but fall short in semantic and realistic aspects (Tabel \ref{tab-summary}). Thus, the CAE decoder can function as the mapping network incorporated into the Map method. In this case, reconstructed brain-conditional images from CAEA can serve as image conditions within the Image-to-Image pipeline, offering low-level detail priors for subsequent conditional generative models.

\begin{table*}[htbp]
  \centering
  \caption{AIGC-Brain Decoding Models for AIGC-Brain Tasks.}
    \renewcommand{\arraystretch}{1.3}
    \resizebox{\textwidth}{!}{
    \begin{threeparttable}
    \begin{tabular}{c|c|c|c|c|c|c|c|c}
    \thickhline
    AIGC Task  & Dataset & Model & Method & Network & Detail & Semantic & Decoder & Type \\
    \hline
    \multirow{38}{*}{IBI} & \multirow{10}{*}{NSD}   & BrainSCN\cite{BrainSCN} & Map   & SCN   & Noise Vector & Instance Features (SwAV) & ICGAN\cite{ICGAN}  & GAN \\
\cline{3-9}          &    & MindReader\cite{Mind-Reader} & Map   & CNN   & -     & CLIP-Text\&Image & StyleGAN2\cite{StyleGAN2} & GAN \\
\cline{3-9} & & BrainSD\cite{BrainSD} & Map   & Ridge & Pixel & CLIP-Text & SD\cite{SD}    & Diffusion \\
\cline{3-9}          &    & BrainSD-TGD\cite{BrainSD-TGD} & Map   & Ridge & Pixel\&Depth & CLIP-Text & SD\cite{SD}    & Diffusion \\
\cline{3-9}          &    & BrainCLIP\cite{BrainClip} & Map   & VAE   &  Pixel (Retrieval) & CLIP-Text\&Image & GD\cite{GD}    & Diffusion \\
\cline{3-9}          &    & BrainDiffuser\cite{BrainDiffuser} & Map   & Ridge & Pixel & CLIP-Text\&Image & VD\cite{VD}    & Diffusion \\
\cline{3-9}          &    & MindDiffuser\cite{MindDiffuser} & Map   & Ridge & Pixel & CLIP-Text\&Image & SD\cite{SD}    & Diffusion \\
\cline{3-9}          &    & UniBrain\cite{UniBrain} & Map   & Ridge & Pixel & CLIP-Text\&Image & VD\cite{VD}    & Diffusion \\
\cline{3-9}          &    & MindEye\cite{MindEye} & Map   & MLP   & Pixel & CLIP-Image & VD\cite{VD}    & Diffusion \\
\cline{3-9}          &    & Dream\cite{Dream} & Map   & CNN\&MLP & Color\&Depth & CLIP-Text\&Image & SD\cite{SD}    & Diffusion \\
\cline{2-9}          & \multirow{4}{*}{GOD/BLOD} & Mind-Vis\cite{Mind-Vis} & BPFA  & MAE/-     & -     & fMRI Features & SD\cite{SD}    & Diffusion \\
\cline{3-9}          &  & CMVDM\cite{CMVDM} & BPFA  & MAE/-     & Silhouette & fMRI Features & SD+ControlNet\cite{ControlNet} & Diffusion \\
\cline{3-9}          &  & CAD\cite{CAD} & BPFA  & DC-MAE/Attention & -     & fMRI\&Visual & SD\cite{SD}    & Diffusion \\
\cline{3-9}          &  & LEA\cite{LEA}   & BPM   & MAE/Ridge & -     & CLIP-Image & MaskGIT\cite{MaskGit} & Autoregressive \\
\cline{2-9}          & \multirow{14}{*}{GOD}   & DBDM\cite{DBDM}  & Map   & MLP   & Pixel & CLIP-Text\&Image & SD\cite{SD}    & Diffusion \\
\cline{3-9}          &   & SBD\cite{SBD}   & Map   & Ridge & -     & Deep-Visual (ResNet) & SD\cite{SD}    & Diffusion \\
\cline{3-9}          &    & BrainICG\cite{BrainICG} & Map   & Ridge & Noise Vector & Instance Features (SwAV) & ICGAN\cite{ICGAN} & GAN \\
\cline{3-9}          &    & BrainHSG\cite{BrainHSG} & Map   & DNN   & Shallow-Visual (ResNet) & Deep-Visual (VGG) & GAN   & GAN \\
\cline{3-9}          &    & BrainHVAE\cite{BrainHVAE}  & Map   & FCN & Shallow-Visual (VGG) & Deep-Visual (VGG) & VAE   & VAE \\
\cline{3-9}          &    & BrainDCG\cite{BrainDCG} & Map   & Ridge & Shallow-Visual (AlexNet) & Deep-Visual (AlexNet) & DCGAN\cite{DCGAN} & GAN \\
\cline{3-9}          &    & DIR\cite{DIR}   & Map   & SLR   & Shallow-Visual (VGG) & Deep-Visual (VGG) & DGN\cite{DGN}   & GAN \\
\cline{3-9}          &    & BrainSSG\cite{BrainSSG} & Map   & FCN\&DNN & Shape & Class Features  & I2I-GAN\cite{I2I-GAN} & GAN \\
\cline{3-9}          &    & BrainBBG\cite{BrainBBG} & Map   & Ridge & -     & Latent Vectors & BigBiGAN\cite{BigBiGAN} & GAN \\
\cline{3-9}          &    &SSNIR\cite{SSNIR} & CAEA & CNN   & -     & fMRI\&Visual & DeCNN & CAE \\
\cline{3-9}          &    &SSNIR-SC\cite{SSNIR-SC} & CAEA & CNN   & -     & fMRI\&Visual & DeCNN & CAE \\
\cline{3-9}          &    &SSDR\cite{SSDR} & CAEA & CNN   & -     & fMRI\&Visual & DeCNN & CAE \\
\cline{3-9}          &    & BrainDVG\cite{BrainDVG} & MTF   & MLP   & -     & Latent Vectors & VAE-GAN & VAE\&GAN \\
\cline{3-9}          &    & E-DIR\cite{E-DIR} & E2E   & -     & -     & fMRI Features & GAN   & GAN \\
\cline{2-9}          & \multirow{7}{*}{EEG-VOA} & DreamDiffusion\cite{DreamDiffusion} & BPFA  & MAE/Projection & -     & EEG\&CLIP-Image & SD\cite{SD}    & Diffusion \\
\cline{3-9}          &  & NeuroImagen\cite{NeuroImagen} & Map   & Extractors & Pixel & CLIP-Text & SD\cite{SD}    & Diffusion \\
\cline{3-9}          &  & DM-RE2I\cite{DM-RE2I} & MTF   & EVRNet & -     & Class Features & DDPM  & Diffusion \\
\cline{3-9}          &  & NeuroVision\cite{NeuroVision} & MTF   & LSTM+FCN & -     & Class Features & C-PGGAN & GAN \\
\cline{3-9}          &  & EEG-VGD\cite{EEG-VGD} & MTF   & CNN   & -     & Class Features & GAN   & GAN \\
\cline{3-9}          &  & EEG-GAN\cite{EEG-GAN} & MTF   & LSTM  & -     & Class Features & GAN   & GAN \\
\cline{3-9}          &  & Brain2Image\cite{Brain2Image} & MTF   & LSTM & -     & Class Features & VAE/GAN & VAE/GAN \\
\cline{2-9}          & MEG-Things & MEG-BD\cite{MEG-BD} & Map   & CNN & Pixel & CLIP-Text\&Image & SD\cite{SD}    & Diffusion \\
    \hline
    \multirow{6}{*}{VBV} & \multirow{5}{*}{DNV} & DNV\cite{DNV}   & CAEA & CNN   & -     & fMRI Features & DeCNN & CAE \\
\cline{3-9}          &  & SSRNM\cite{SSRNM} & CAEA & CNN   & -     & fMRI\&Visual & DeCNN & CAE \\ 
\cline{3-9}          &  & BrainViVAE\cite{BrainViVAE} & Map   & LR & -     & Latent Vectors & VAE   & VAE\\
\cline{3-9}          &  & f-CVGAN\cite{f-CVGAN} & E2E   & FCN & -     & fMRI Features & GAN   & GAN \\
\cline{3-9}          &  & Mind-Video\cite{Mind-Video} & BPFA  & MAE/Contrastive & -     & fMRI\&CLIP & Augmented SD & Diffusion \\

\cline{2-9}          & STNS & Brain2Pix\cite{Brain2Pix} & E2E   & -     & -     & fMRI Features & GAN   & GAN \\
    \hline
    \multirow{2}{*}{SBS} & BSR  & BSR\cite{BSR} & Map   & Ridge & -     & Deep-Sound (DNN) & Audio-Transformer & Autoregressive \\
\cline{2-9}          & ETCAS & ETCAS\cite{ETCAS} & E2E   & -     & -     & EEG Features & Dual-DualGAN & GAN \\
    \hline
    \multirow{2}{*}{MBM} & MusicGenre & Brain2Music\cite{Brain2Music} & Map   & Ridge & w2v-BERT Tokens & MuLan Embedding & MusicLM\cite{musiclm} & Autoregressive \\
\cline{2-9}          & MusicAffect & NDMusic\cite{NDMusic} & E2E   & -     & -     & fMRI-EEG Features & BiLSTM & Autoregressive \\
    \hline
    \multirow{7}{*}{IBT} & \multirow{3}{*}{NSD}   & UniBrain\cite{UniBrain} & Map   & Ridge & Word\&Sentence & CLIP-Text\&Image & VD\cite{VD}    & Diffusion \\
\cline{3-9}          &   & BrainCaptioning\cite{BrainCaptioning} & Map   & Ridge & -     & CLIP-Image & GIT\cite{GIT}   & Autoregressive \\
\cline{3-9}          &   & DreamCatcher\cite{DreamCatcher} & Map   & FCN & -     & GPT-Embedding & LSTM  & Autoregressive \\
\cline{2-9}          & \multirow{2}{*}{OCD}   & GIC-PTL\cite{GIC-PTL} & MTF   & BiGRU & -     & Deep-Visual (CNN) & GRU   & Autoregressive \\
\cline{3-9}          &    & GIC-CT\cite{GIC-CT} & E2E   & CNN & -     & fMRI Features & Transformer & Autoregressive \\
\cline{2-9}          & \multirow{2}{*}{VER} & DSR\cite{DSR}   & Map   & FCN & -     & Deep-Visual (VGG) & LSTM  & Autoregressive \\
\cline{3-9}          &  & GNLD\cite{GNLD}   & Map   & MLP & -     & Deep-Visual (VGG) & LSTM  & Autoregressive \\
    \hline
    SBT   & Narratives & UniCoRN\cite{UniCoRN} & BPM   & CAE/Projection & -     & BART-Embedding & BART\cite{bart}  & Autoregressive \\
    \hline
    SBT\&VBT & CLSR  & CLSR\cite{CLSR}  & Map   & Ridge & -     & GPT-Embedding & GPT\cite{gpt}   & Autoregressive \\
    \thickhline
    \end{tabular}%
    \begin{tablenotes} 
        \item[1] The term 'Network' refers to corresponding networks utilized by different methods corresponding to Fig.\ref{fig_method}, that is, the mapping network (MapNet) in Map and MTF methods (separate different detail and semantic mapping networks by `$\&$'), the feature network (FeatNet) in the E2E method, the CAE encoder in the CAEA method, and the brain encoder with mapping/alignment network in BPM and BPFA methods (denoted as `\textit{brain-encoder/network}'), respectively.
    \end{tablenotes} 
    \end{threeparttable}
  \label{tab-work}%
  }
\end{table*}%

\section{AIGC-Brain Tasks\&Implementations}
\label{sec: tasks}
In this section, we delve into the intricacies of each AIGC-Brain task, offering a detailed analysis that integrates AIGC-Brain implementation strategies. The recent advancements in AIGC-Brain, spanning various tasks, are succinctly summarized in Table \ref{tab-work}. This table systematically organizes each model based on its specific task and dataset, providing details on the employed methods (outlined in Section \ref{sec:method}), networks, detail priors, semantic priors, AIGC decoders, and decoder types. 
Here, the term 'Network' refers to corresponding networks utilized by different methods, that is, the mapping network (MapNet) in Map and MTF methods, the feature network (FeatNet) in the E2E method, the CAE encoder in the CAEA method, and the brain encoder with the mapping/alignment network (represented as encoder\&network) in BPM and BPFA methods, respectively. 

For a comprehensive overview, we primarily focus on introducing pertinent work associated with representative datasets for each task in Table \ref{tab-work}. Unmentioned datasets and their related work are cataloged in the 'task' column of Table \ref{tab-datasets}, and implementation details can be found in their respective original papers to which we kindly direct the reader for a more in-depth understanding.

\subsection{Image-Brain-Image}
\label{subsec: IBI}
The Image-brain-Image (IBI) task has garnered significant attention in the current AIGC-Brain domain, with a considerable body of research. In this session, we primarily consolidate the efforts undertaken on the fMRI-Image datasets (NSD and GOD), the EEG-Image dataset (EEG-VOA), and the MEG-Image dataset (MEG-Things).

Image-to-image latent diffusion models (I2I-LDMs), incorporating pixel detail priors and CLIP semantic conditional priors, stand out as the current state-of-the-art models across all representative datasets for the IBI task (NSD: BrainSD, BrainDiffuser, UniBrain, MindEye; GOD: DBDM; EEG-VOA: NeuroImagen; MEG-Things: MEG-BD). The framework of brain-conditional I2I-LDMs based on the Map method is illustrated in Fig.\ref{fig_IBI_map}. Specifically, brain signals evoked by image stimuli are first mapped to pixel priors and CLIP semantic priors. Following this, the pixel priors are decoded into pixel images, serving as the initial guess. The I2I-LDM decoder further involves the forward noise-adding process and CLIP-guided backward denoising processes, ultimately culminating in the production of high-quality images. Note that CLIP semantic priors consist of CLIP-Text and/or CLIP-Image features derived from the pretrained text and image encoders of CLIP\cite{Clip}.

\subsubsection{fMRI-Based Models}
\begin{figure}[!t]
\centering
\includegraphics[width=\linewidth]{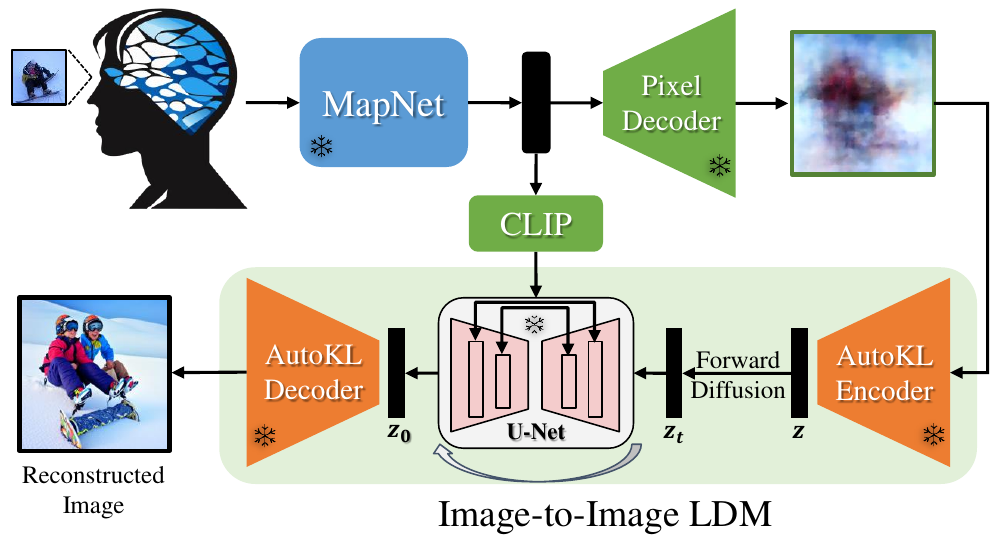}
\caption{Brain-Conditional I2I-LDMs: State-of-the-art framework for IBI tasks.} 
\label{fig_IBI_map}
\end{figure}

\paragraph{NSD-Based}
Upon the NSD dataset, all decoding models leverage the Map method with mapping networks and pretrained AIGC decoders including Diffusion and GAN. Linear regressions (i.e., Ridge regression and sparse linear regression (SLR))  are commonly used as mapping networks between fMRI voxels and prior information. BrainSCN\cite{BrainSCN} introduces a surface-based convolutional neural network (SCN) to map fMRI voxels to noise vectors and instance features extracted from pretrained SwAV\cite{Swav}, which are then decoded into images by the generator of pretrained ICGAN\cite{ICGAN}. In the case of MindReader\cite{Mind-Reader}, a typical CNN and StyleGAN2\cite{StyleGAN2} are employed without incorporating detail priors. Notably, BrainSD pioneers the utilization of Ridge regression and the I2I pipeline with Stable Diffusion (SD) for IBI tasks (Fig.\ref{fig_IBI_map}). Subsequent studies are built upon this foundation, making enhancements such as replacing simple linear regression with non-linear neural networks (e.g., CNN, VAE, multilayer perception (MLP)), incorporating additional details like depth and color, introducing dual conditions from CLIP (CLIP-Text\&Image) to enhance semantic fidelity, and using more powerful diffusion models (e.g., Versatile Diffusion (VD)\cite{VD}, Guided Diffusion (GD)\cite{GD}). These changes are easy to implement as mapping networks are lightweight and work independently. The most important thing is to select an appropriate pretrained AIGC decoder to combine the mapped prior information of each part for image generation.

\paragraph{GOD-Based}
Upon the GOD dataset, AIGC-Brain decoding models mainly leverage the Map, BPFA, and AEA methods. In terms of the Map method, DBDM is the state-of-the-art model that utilizes brain-conditional I2I-LDMs with a mapping network of MLP, as depicted in Fig.\ref{fig_IBI_map}. BrainICG resembles BrainSCN except for the mapping networks (SCN vs. Ridge). BrainSSG utilizes a fully connected network (FCN) and deep neural network (DNN) to map fMRI voxels to shape detail priors and semantic class features from image labels, respectively. Moreover, brain signals can be mapped to hierarchy visual features extracted from pretrained deep neural networks (e.g., ResNet\cite{ResNet}, VGG\cite{VGG}, AlexNet\cite{AlexNet}) or latent vectors extracted from pretrained AIGC decoders (e.g., BigBiGAN\cite{BigBiGAN}, VAE\cite{VAE}). Hierarchy visual features encompass detail features from shallow layers (Shallow-Visual) and semantic features from deep layers (Deep-Visual), which can be further decoded into images by pretrained AIGC decoders (e.g., VAE, DCGAN\cite{DCGAN}, and deep generator network (DGN)\cite{DGN}.

The BPFA method, along with Diffusion decoders, is currently the state-of-the-art method when conducting the IBI task on GOD, BLOD, and the EEG-VOA datasets. As a pioneer work, Mind-Vis employs Masked Autoencoders (MAE)\cite{MAE} for brain pretraining on the large-scale unlabeled fMRI dataset HCP\cite{HCP}. MAE enables the model to learn meaningful representations by partially hiding and reconstructing the input data in the pretraining stage. After that, the brain encoder and the attention layer of the pretrained SD model are finetuned to generate brain-conditional images. Mind-Vis\cite{Mind-Vis} open-sources the pretrained weights that enable subsequent work to improve upon them. CVMDM\cite{CMVDM} designs an extra brain-to-silhouette decoder for detail priors and combines them with SD using the idea of ControlNet \cite{ControlNet} in the finetuning stage, thus improving the generation performance of Mind-Vis. CAD\cite{CAD} performs brain pretraining on the HCP dataset using Double-Contrastive MAE (DC-MAE) inspired by previous work in visual contrastive learning\cite{VCL}. DC-MAE helps identify common patterns of brain activity across populations rather than individual variations.
Furthermore, CAD designs a cross-attention layer to perform cross-modal alignment between brain and visual information during finetuning. The BPM-based method LEA\cite{LEA} employs MAE for brain pretraining and adopts Ridge regression to map brain latent features to CLIP-Image semantic priors. Unlike other AIGC-Brain decoders, LEA utilizes the autoregressive-based Masked Generative Image Transformer (MaskGIT)\cite{MaskGit} for image generation.
The BPFA method is limited by its characteristics of hard-pretraining and hard-finetuning. A critical constraint is the necessity for proximity between the pretraining and fine-tuning datasets, introducing challenges when dealing with divergent data types (e.g., fMRI vs. EEG) and varying data lengths (e.g., HCP vs. NSD). Consequently, the adoption of this methodology is hindered, with scant utilization observed in the literature. Researchers predominantly rely on pretrained weights sourced from Mind-Vis, trained on the HCP dataset, and apply them to downstream datasets such as GOD and BOLD. These challenges underscore the necessity for more adaptive approaches in pretraining, especially when confronted with the inherent heterogeneity across neuroscientific datasets.

Furthermore, SSNIR, SSNIR-SC, and SSDR employ the CAEA method with modality alignment using supervised learning (SL) and self-supervised learning (SSL) for IBI tasks. They mainly consist of three stages: i) Image-to-Brain CAE encoder training (SL), ii) Brain-to-Image CAE decoder training (SL), and iii) Image-to-Image CAE decoder training (SSL). However, their generated images tend to be more inclined to low-level details but lack semantics and authenticity. To tackle this problem, generated images from CAEA can serve as image conditions within the Image-to-Image pipeline, offering low-level detail priors for subsequent conditional generative models. Specifically, Dream\cite{Dream},  the Map-based model, adopts SSDR\cite{SSDR} as the mapping network to map brain signals to color and depth images for detail priors.

\subsubsection{EEG-Based Models}
The MTF method is mainly used for EEG-based IBI tasks. Unlike fMRI voxels in one dimension, EEG signals are typically two-dimensional data involving time and channels. Therefore, typical mapping networks in MTF are neural networks like LSTM, CNN, GRU, MLP, and FCN. Researchers first establish the relationship between EEG signals and class labels, high-level visual information, or latent vectors via the mapping network. Then, the mapped semantic features are transmitted to the AIGC decoder for finetuning or training from scratch. Specifically, DM-RE2I\cite{DM-RE2I} introduces a well-designed residual network EVRNet for class features and trains a denoising diffusion probabilistic model (DDPM)\cite{DDPM} for image synthesis. NeuroVision\cite{NeuroVision} combines LSTM and FCN for feature extraction and trains a class-conditional PGGAN (C-PGGAN) for image generation. Recently, based on the BPFA method, DreamDiffusion\cite{DreamDiffusion} reaches state-of-the-art performance in EEG-based IBI tasks following the idea of Mind-Vis. DreamDiffusion collects a large-scale unlabeled EEG dataset on the MOABB\cite{MOABB} platform and performs brain pretraining on it using the MAE method. In the finetuning stage, DreamDiffusion follows the basic idea of Mind-Vis and employs a projection layer to align brain latent representation with CLIP-Image semantic information. DreamDiffusion employs a similar approach to Mind-Vis in the EEG domain, but its pretraining dataset and weights have not been released so far. Based on the Map method, NeuroImagen employs detail and semantic extractors to map EEG signals to pixel and CLIP-Text priors, which are then decoded by pretrained SD following the image-to-image pipeline (Fig.\ref{fig_IBI_map}).

\subsubsection{MEG-Based Models}
MEG-BD\cite{MEG-BD}, the sole existing investigation on the MEG-Things dataset, employs the Map methodology that leverages CNN as the mapping network and I2I-LDM as the AIGC-Brain decoder, as illustrated in Fig.\ref{fig_IBI_map}. Diverging from the prevalent usage of fMRI and EEG modalities, MEG uniquely ensures temporal and spatial resolution concurrently. This pioneering work lays the foundation for the exploration of the MEG-based IBI tasks.

\subsection{Video-Brain-Video}
\label{subsec: VBV}
\begin{figure}[!t]
\centering
\includegraphics[width=\linewidth]{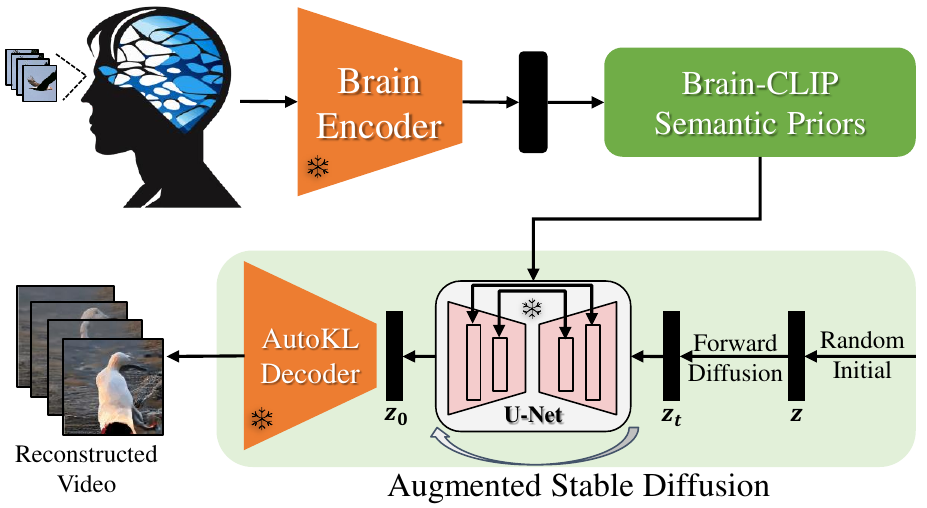}
\caption{Mind-Video\cite{Mind-Video}: State-of-the-art model for VBV tasks.} 
\label{fig_VBV_map}
\end{figure}

Video-Brain-Video (VBV) tasks are addressed using Map, BPFA, and E2E decoding methods. BrainViVAE\cite{BrainViVAE} employs linear regression to map brain signals to latent vectors of the pretrained VAE, which are then decoded into videos. f-CVGAN\cite{f-CVGAN} and Brain2Pix\cite{Brain2Pix} utilize end-to-end GAN models that directly connect brain signals to corresponding videos without relying on prior information. SSRNM\cite{SSRNM} follows the idea of CAEA-based image-generated models (i.e., SSNIR, SSNIR-SC, and SSDR) and extends it for VBV tasks. Additionally, Wen et al. \cite{DNV} employ a simple version of CAEA that relies solely on supervised learning for modality alignment. Building on the BPFA method, Mind-Video\cite{Mind-Video} extends the work of Mind-Vis in the VBV domain. Mind-Video leverages the pretrained brain encoder from Mind-Vis for brain latent features, aligning them with CLIP-Image and CLIP-Text features using contrastive learning. The video generation process incorporates an augmented stable diffusion model with spatial and temporal attention. During the finetuning stage, adversarial semantic features derived from fMRI act as guidance for video generation. Mind-Video serves as the state-of-the-art model for VBV tasks and its inference process is depicted in Fig.\ref{fig_VBV_map}.

\subsection{Sound-Brain-Sound}
\label{subsec: SBS}

\begin{figure}[!t]
\centering
\includegraphics[width=\linewidth]{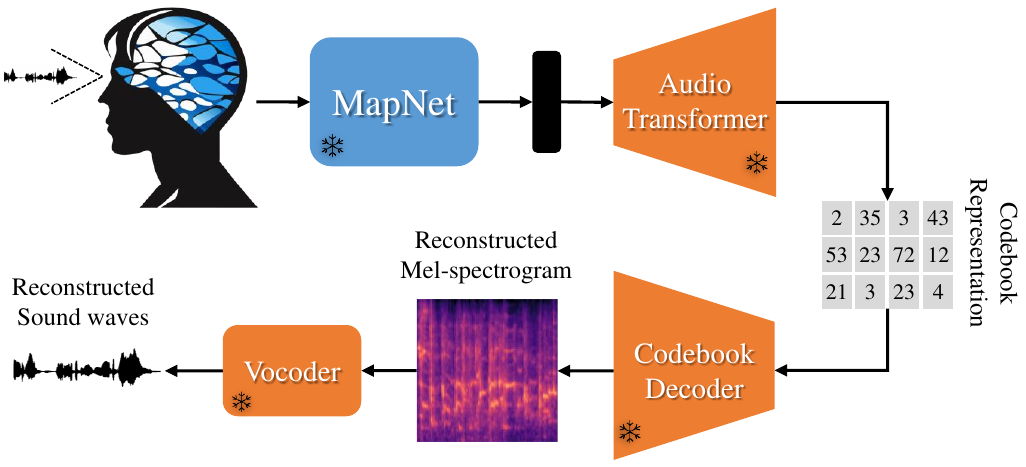}
\caption{BSR\cite{BSR}: State-of-the-art model for SBS tasks.} 
\label{fig_SBS_map}
\end{figure}

In the realm of SBS tasks, our investigation has identified two relevant works grounded in the Map and E2E methodologies. ETCAS\cite{ETCAS}, an end-to-end GAN model tailored for EEG-based SBS tasks, introduces a Dual-DualGAN to directly map EEG signals to speech signals. In particular, BSR\cite{BSR}, adopting the Map method with an autoregressive decoder, employs Ridge regression to map fMRI voxels to semantic sound features. These Deep-Sound features are extracted from deep layers of a pretrained sound recognition DNN, which is performed on Mel-spectrograms of natural sound stimuli. The mapped semantic features are then fed into a pretrained audio transformer to generate codebook representation. The pretrained codebook decoder receives this representation and reproduces the Mel-spectrogram, subsequently converted into sound waves by a pretrained Vocoder, as depicted in Fig.\ref{fig_SBS_map}.

\begin{figure}[!t]
\centering
\includegraphics[width=\linewidth]{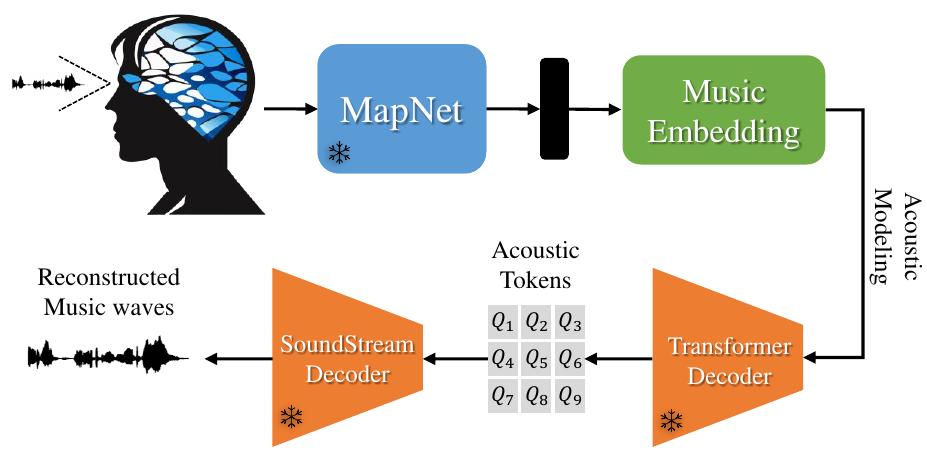}
\caption{Brain2Music\cite{Brain2Music}: State-of-the-art model for MBM tasks.} 
\label{fig_MBM_map}
\end{figure}

\begin{figure}[!t]
\centering
\includegraphics[width=\linewidth]{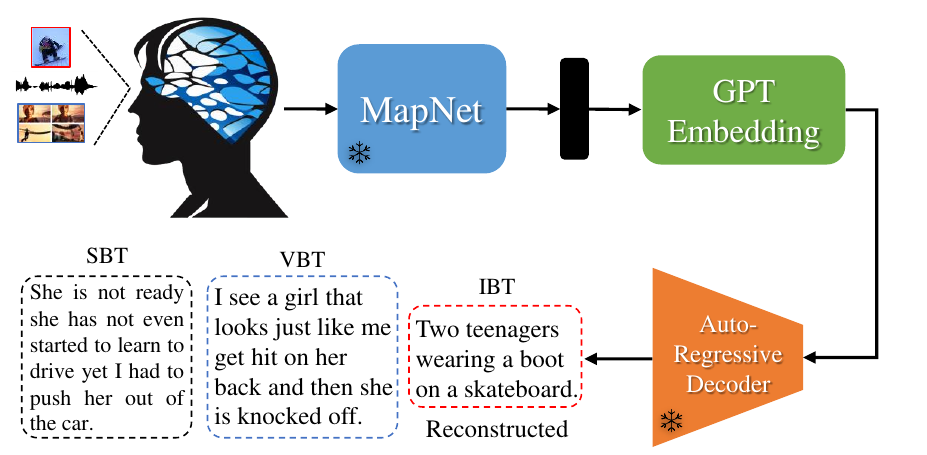}
\caption{Representative framework for IBT, VBT, and SBT tasks.} 
\label{fig_IVSBT_map}
\end{figure}

\subsection{Music-Brain-Music}
\label{subsec: MBM}
For the MBM task, we have identified studies employing the Map and E2E methods. NDMusic adopts an end-to-end bidirectional LSTM (BiLSTM) architecture to establish a direct mapping from fMRI-informed EEG signals to music signals. In particular, Brain2Music follows the Map method, incorporating a mapping network based on Ridge regression and a pretrained AIGC decoder MusicLM\cite{musiclm}. The MusicLM model comprises three key components: i) w2v-BERT\cite{w2v-bert} for speech representation; ii) MuLan\cite{mulan} for contrastive semantic representation in both audio and text domains; and iii) SoundStream \cite{soundstream}, a neural audio codec, responsible for the generation of audio signals. Specifically, as illustrated in Fig.\ref{fig_MBM_map}, brain signals are mapped to music embedding (i.e., w2v-BERT tokens and MuLan embeddings) and decoded into acoustic tokens (latent representation in SoundStream with residual vector quantizer), which are sequentially decoded into music waves via the pretrained SoundStream decoder.

\subsection{Image\&Video\&Speech-Brain-Text}
\label{subsec: IVSBT}
The text generation task involves Map, BPM, MTF, and E2E methods with autoregressive decoders, except that UniBrain employs the diffusion decoder for both IBI and IBT tasks. 

For the IBT task, BrainCaptioning maps fMRI voxels to CLIP-Image features using Ridge regression, which are then decoded into text by the Generative Image-to-text Transformer (GIT) \cite{GIT}. DreamCatcher\cite{DreamCatcher} utilizes FCN to map fMRI voxels to GPT embeddings and employs LSTM as an autoregressive decoder for text generation, as depicted in Fig.\ref{fig_IVSBT_map}. DSR\cite{DSR} and GNLD\cite{GNLD} employ FCN and MLP to map fMRI voxels to visual features extracted from deep layers of VGG, respectively. Both of them use the autoregressive decoder LSTM for text generation. UniBrain\cite{UniBrain} maps fMRI signals to low-level text detail priors (i.e., words and sentences), as well as semantic priors derived from CLIP-Text and CLIP-Image. The pretrained versatile diffusion (VD) decoder receives prior information and generates descriptive captions for image stimuli.

For the SBT task, UniCoRN\cite{UniCoRN} first performs brain pretraining using CAE to extract brain latent features, which are then mapped to the embedding of BART\cite{bart} via a projection network. Finally, BART embeddings are decoded into text by the pretrained BART decoder.
CLSR\cite{CLSR} maps fMRI signals to the GPT embedding via Ridge regression and then decodes it into text by a pretrained GPT, as depicted in Fig.\ref{fig_IVSBT_map}. Furthermore, language features from this GPT embedding can transfer to the VBT task that helps to generate captions for silent video stimuli \cite{CLSR, VL-Transfer}.

\subsection{Other Tasks}
This survey mainly focuses on passive non-invasive AIGC-Brain tasks evoked by sensory stimuli (i.e., vision and audio) as described above, other AIGC-Brain generative tasks are briefly introduced as follows.

\textbf{Active Tasks: }
In contrast to passive tasks, active AIGC-Brain tasks do not necessitate stimulus intervention during the inference stage, leveraging spontaneous brain signals to guide cross-modal content synthesis. This distinctive feature distinguishes them from passive tasks, showcasing the potential for self-driven cognitive processes in shaping multimodal information integration. Presently, we delineate two types of active brain-conditional tasks: i) Reading and Mental Rehearsal: Brain signals induced via reading text are utilized to synthesize semantic textual content, and participants are only required to actively rehearse the text mentally during the inference stage, termed Brain-to-Text\cite{duan2023dewave}; ii) Imagery: Brain signals induced via the imagination of visual scenes or auditory content are utilized to synthesize multimodal content, and participants are only required to actively imagine relevant information during the inference stage. Associated generative tasks conditioned on imagery brain signals involve Brain-to-Image, Brain-to-Speech\cite{CLSR}, and so on. For instance, Brain-to-Image refers to the ability to mentally generate or recall visual experiences without external sensory input. Several studies \cite{ThoughtViz, NeuroGAN, EEG2Image} have implemented EEG-based Brain-to-Image tasks on the EEG-Imagey dataset\cite{EEG-Imagery}, which is collected from 23 subjects imaging three different categories (characters, digits, and objects).

\textbf{Invasive Tasks: }
While invasive techniques often provide more direct access to neural activity, they come with higher risks and ethical considerations due to the need for surgery. Currently, invasive neuroimaging techniques like ECoG (Electrocorticography) and intracranial EEG (iEEG) have been utilized for invasive AIGC-Brain tasks, involving Brain-to-Speech\cite{Ecog-speech}, Brain-to-Music\cite{iEEG-music}, and so on.

\begin{figure*}[!t]
\centering
\includegraphics[width=\linewidth]{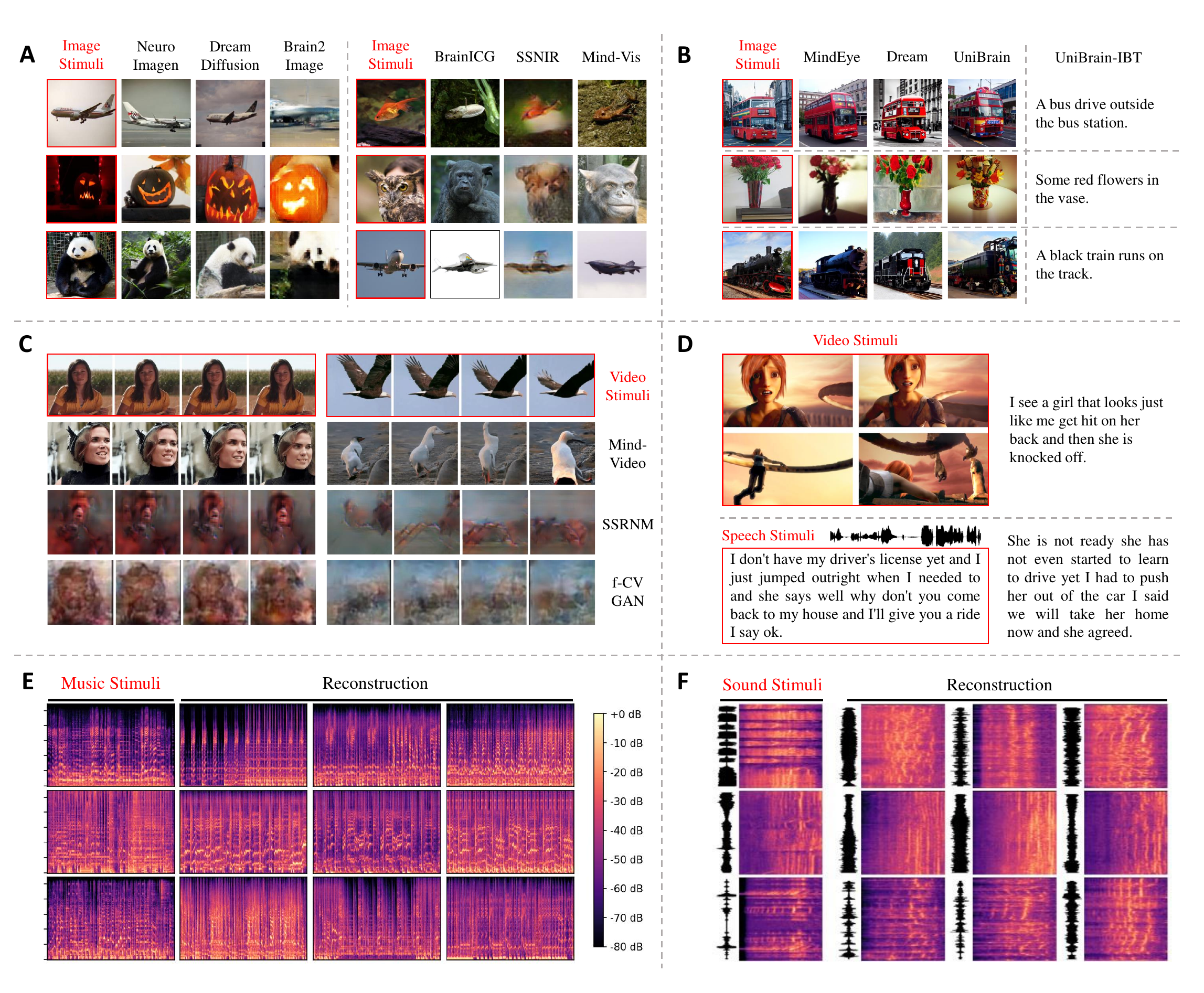}
\caption{Qualitative Results for AIGC-Brain Tasks. A: IBI results on GOD (left) and EEG-VOA (right) datasets; B: IBI and IBT results on NSD dataset; C: VBV results on DNV dataset; D: VBT (top) and SBT (bottom) results based on CLSR\cite{CLSR}; E: MBM results based on Brain2Music\cite{Brain2Music}; F: SBS results based on BSR\cite{BSR}.} 
\label{fig_result}
\end{figure*}

\section{AIGC-Brain Results}
\label{sec: results}

The purpose of AIGC-Brain tasks is to reconstruct the original stimulus or its textual description from the evoked brain signals. To this end, AIGC-Brain results need to focus on the fidelity between the ground truth and generated content, including low-level detail fidelity and high-level semantic fidelity. In this session, we will present qualitative results and quantitative metrics/results that have been summarized from representative datasets and models used for AIGC-Brain tasks. To be fair, we only provide performance comparisons for multiple works with the same dataset as well as the same experimental setup.

\subsection{Qualitative Results}
\label{subsec: result_qualitative}

The qualitative results of representative models for all AIGC-Brain tasks are depicted in Fig.\ref{fig_result}. Generative content is selected from representative models and presented beside corresponding sensory stimuli.
Specifically, Fig.\ref{fig_result}-A focuses on IBI qualitative results on GOD (left) and EEG-VOA (right) datasets. Image stimuli in both GOD and EEG-VOA are subsets from ImageNet. For EEG-VOA, all these models reproduce high semantic fidelity (i.e., panda, Jack-o-lantern, and airplane) and similar details in terms of object color and contour. However, generative results for GOD exhibit inadequate semantic or detail fidelity. For instance, SSNIR generates images with high detail fidelity but is burry. BrainICG and Mind-Vis generate realistic images but semantic mismatch (i.e., object category) and detail mismatch (i.e., object color) still exist. This phenomenon is mainly caused by the difficulty of zero-shot tests on GOD. 

Fig.\ref{fig_result}-B focuses on IBI (left) and IBT (right) qualitative results on the NSD dataset. All these state-of-the-art models reproduce images with both high semantic (i.e., object relationship, scene understanding) and detail fidelity (i.e., pixel and contour). UniBrain \cite{UniBrain} generates high-fidelity images and well-described captions from image-evoked brain signals, respectively.

Fig.\ref{fig_result}-C focuses on VBV qualitative results on the DNV dataset. SSRNM\cite{SSRNM} and f-CVGAN\cite{f-CVGAN} only reconstruct blurred video frames with rough object contours, which are not realistic enough. Mind-Video produces high-quality video frames while maintaining similar semantic information in terms of object category (i.e., woman and pigeon). However, the fidelity of details (i.e., pixel, contour, and motion vectors) still needs to be improved.

Fig.\ref{fig_result}-D focuses on VBT (top) and SBT (bottom) qualitative results based on the dataset and model from CLSR\cite{CLSR}. CLSR generates textual descriptions from brain signals evoked by video and speech stimuli. For VBT, the generated captions match the video scenes (i.e., object, motion) and connect them via the timeline (i.e., look, hit, and knock-off). For SBT, the generated language resembles the original speech content in semantics, even though there are differences in low-level text information (i.e., word and phrase).

Fig.\ref{fig_result}-E focuses on MBM qualitative results from Brain2Music\cite{Brain2Music} upon the MusicGenre dataset. The music stimuli and reconstruction samples are shown in the Mel-Spectrogram format. As can be seen, the generated Mel-Spectrograms exhibit similar spectral patterns to the original Mel-Spectrogram. Audio of more original and reconstructed music can be found in this project: \url{https://google-research.github.io/seanet/brain2music/}.

\begin{table*}[htbp]
  \centering
  \caption{Quantitative Results of AIGC-Brain Models on the \textbf{NSD} Dataset for \textbf{IBI}.}
    \renewcommand{\arraystretch}{1.1}
    \resizebox{\textwidth}{!}{
    \begin{tabular}{ccccclcccc}
    \thickhline
    \multirow{2}[1]{*}{Model} & \multicolumn{4}{c}{Low-Level} & & \multicolumn{4}{c}{High-Level} \\
    \cline{2-5} \cline{7-10} & PixCorr \textbf{$\uparrow$} & SSIM\textbf{ $\uparrow$} & AlexNet-2\textbf{ $\uparrow$} & AlexNet-5\textbf{ $\uparrow$} & & Inception\textbf{ $\uparrow$} & CLIP\textbf{ $\uparrow$} & EffNet\textbf{ $\downarrow$} & SwAV\textbf{ $\downarrow$} \\
    \hline
    Mind-Reader\cite{Mind-Reader} & -     & -     & -     & -   &  & 0.782 & -     & -     & - \\
    BrainSCN\cite{BrainSCN} & 0.15  & 0.325 & -     & -  &   & -     & -     & 0.862 & 0.465 \\
    BrainSD\cite{BrainSD} & -     & -     & 0.83  & 0.83 & & 0.76  & 0.77  & -     & - \\
    BrainSD-TGD\cite{BrainSD-TGD} & -     & -     & -     & -  &   & 0.843 & 0.866 & -     & - \\
    MindDiffuser\cite{MindDiffuser} & -     & 0.354 & -     & -  &   & -     & -     & -     & - \\
    BrainDiffuser\cite{BrainDiffuser} & 0.254 & 0.356 & 0.942 & 0.962 & & 0.872 & 0.915 & 0.775 & 0.423 \\
    BrainCLIP\cite{BrainClip} & -     & -     & -     & -  &   & 0.867 & 0.948 & -     & - \\
    UniBrain\cite{UniBrain} & 0.249 & 0.330 & 0.929 & 0.956 & & 0.878 & 0.923 & 0.766 & 0.407 \\ 
    MindEye\cite{MindEye} & 0.309 & 0.323 & 0.947 & 0.978 & & 0.938 & 0.941 & 0.645 & 0.367 \\
    Dream\cite{Dream} & 0.288 & 0.338 & 0.95  & 0.975 & & 0.948 & 0.952 & 0.638 & 0.413 \\
    \thickhline
    \end{tabular}%
    }
  \label{tab_resultNSD}%
\end{table*}%

\begin{table*}[htbp]
  \centering
  \caption{Quantitative Results of AIGC-Brain Models on the \textbf{GOD} Dataset for \textbf{IBI}.}
    \renewcommand{\arraystretch}{1.1}
    \resizebox{0.91\textwidth}{!}{
    \begin{tabular}{cccclcccc}
    \thickhline
    \multirow{2}[1]{*}{Model} & \multicolumn{3}{c}{Low-Level} & & \multicolumn{4}{c}{High-Level} \\
    \cline{2-4} \cline{6-9} & PixCorr\textbf{ $\uparrow$} & SSIM\textbf{ $\uparrow$} & PCC\textbf{ $\uparrow$} & & Inception-Dist\textbf{ $\downarrow$} & CLIP-Dist\textbf{ $\downarrow$} & SwAV\textbf{ $\downarrow$} & ACC\textbf{ $\uparrow$} \\
    \hline
    DIR\cite{DIR}   & 0.339 & 0.539 & -  &   & 0.933 & 0.379 & 0.581 & - \\
    SSNIR\cite{SSNIR} & 0.351 & 0.575 & 0.482 & & 0.896 & 0.415 & 0.690 & 4.288 \\
    SSNIR-SC\cite{SSNIR-SC} & 0.459 & 0.607 & 0.683 & & 0.871 & 0.389 & 0.592 & 9.128 \\
    BrainBBG\cite{BrainBBG} & 0.103 & 0.431 & -  &   & 0.932 & 0.346 & 0.577 & - \\
    BrainDVG\cite{BrainDVG}   & 0.657 & 0.605 & -   &  & 0.838 & 0.393 & 0.617 & - \\
    BrainICG\cite{BrainICG} & 0.223 & 0.453 & 0.449 & & 0.846 & 0.340 & 0.510 & 29.386 \\
    BrainCLIP\cite{BrainClip} & 0.175 & 0.448 &- &  & 0.908 & 0.301 & 0.527 & - \\
    DBDM\cite{DBDM}  & 0.231 & 0.473 & -  &   & 0.611 & 0.225 & 0.405 & - \\
    Mind-Vis\cite{Mind-Vis} & -     & 0.527 & 0.532 &  & -     & -     & -     & 26.644 \\
    CMVDM\cite{CMVDM} & -     & 0.632 & 0.768 & & -     & -     & -     & 30.112 \\
    \thickhline
    \end{tabular}%
    }
  \label{tab_resultGOD}%
\end{table*}%

Fig.\ref{fig_result}-F focuses on SBS qualitative results based on the dataset and model from BSR\cite{BSR}. The sound stimuli and reconstruction samples are shown in the waveform and Mel-Spectrogram formats. From top to bottom, sound stimuli are from chicken clucking, cow lowing, and donkey, respectively. As can be seen, some of the generated speech signals and Mel-Spectrograms exhibit similar waveform and spectral patterns to the original Mel-Spectrogram.

\subsection{Quantitative Results}
\label{subsec: result_quantitative}
To quantitatively compare current models, we summarize various visual, auditory, and textual evaluation metrics in both low-level and high-level aspects that have been utilized in previous work. 

\subsubsection{Image Metric}
\label{subsubsec: vis_metric}

In this session, we summarize image reconstruction metrics in low-level and high-level aspects performed on the EEG-image dataset (EEG-VOA) and two representative fMRI-image datasets (NSD and GOD). 

\textbf{Low-level:} Low-level image features provide basic information about the visual content and structure of the image, measured by: i) {\textbf{PixCorr:}} Pixel-level correlation of reconstructed and ground-truth images; ii) {\textbf{SSIM:}} Structural Similarity Index \cite{SSIM}; iii) MSE: Mean square error; iv) {\textbf{AlexNet:}} AlexNet-2 and AlexNet-5 are the 2-way comparisons of the second (early) and fifth (middle) layers of AlexNet \cite{AlexNet}, respectively; v) \textbf{PCC:} The Pearson Correlation Coefficient measures the strength of the linear relationship between the pixel values of the reconstructed image and the original image.  

\textbf{High-level:} High-level image features capture semantic information, object relationships, and contextual understanding of the image, measured by: i){\textbf{Inception:}} A two-way comparison of the last pooling layer of InceptionV3 \cite{Inception}; ii) {\textbf{CLIP:}} A two-way comparison of the output layer of the CLIP-Image \cite{Clip} model; iii) {\textbf{EffNet, SwAV, Inception-Dist, CLIP-Dist:}} Distance metrics gathered from EfficientNet-B1 \cite{EffNet}, SwAV-ResNet50 \cite{Swav}, InceptionV3\cite{Inception} and CLIP-Image\cite{Clip} models, respectively; iv) \textbf{IS}: Inception score measures the realism and diversity of the generated images; v) \textbf{IC}: Inception Classification Accuracy refers to the accuracy of a generative model in producing images that the pretrained Inception-v3 classification model correctly classifies; vi) \textbf{ACC}: The 50-way top-1 accuracy that measures the proportion of correctly classified samples from 50 possible classes, where the predicted class is the top prediction. 

The quantitative results of AIGC-Brain models on NSD, GOD, and EEG-VOA datasets for the IBI task are presented in Table \ref{tab_resultNSD}, \ref{tab_resultGOD}, and \ref{tab_resultEEG}, respectively. The up-arrow in the tables means higher is better, and the down-arrow is vice versa. 

\subsubsection{Video Metric}
\label{subsubsec: vid_metric}
The main difference between video features and image features lies in the temporal aspect. Video features encompass both spatial and temporal information, capturing changes over time through elements like motion vectors and optical flow. We summarize video reconstruction metrics in low-level and high-level aspects performed on the representative fMRI-Video dataset DNV\cite{DNV} in Table \ref{tab_resultVid}. Similar to image metrics, SSIM, MSE, and Peak Signal-to-Noise Ratio (PSNR) are used as low-level metrics, and ACC is used as high-level metrics. Furthermore, the Rank score is determined by computing the similarity of short reconstructed clips to n clips (including one correct and n-1 distractors) from the real test video, sorting the computed similarities, and assigning a score based on the position of the correct ground-truth clip within the sorted list, with a lower score indicating higher similarity \cite{SSRNM}. However, these metrics are mainly focused on frame-based spatial and object category consistency, and sequence-based temporal consistency should be further considered. Temporal consistency among frames could be measured by Temporal Structural Similarity (t-SSIM), Temporal-Geometric Consistency (TGC), and Temporal-Geometric Consistency (TGC), etc.

\subsubsection{Sound\&Speech Metric}
\label{subsubsec: speech_metric}
For sound reconstruction, the fidelity and quality of BSR\cite{BSR} are assessed via an identification analysis that incorporates pixels of the Mel-spectrogram, hierarchical sound representation of DNN layers, and acoustic features like the fundamental frequency (F0), the spectral centroid (SC), and the harmonic to noise ratio (HNR). Experimental results demonstrated that the BSR model can not only reconstruct approximate spectral patterns but also the perceptual qualities akin to actual sound stimuli.

For speech reconstruction, ETCAS\cite{ETCAS} measures the generative performance by ACC, PCC\cite{PCC}, and Mel-cepstral distance (MCD)\cite{Mel-cepstral}. Among them, PCC and MCD are low-level speech metrics, while the ACC serves as the high-level metric.

\begin{table}[htbp]
  \centering
  \caption{Quantitative Results of AIGC-Brain Models on the \textbf{EEG-VOA} Dataset for \textbf{IBI}.}
    \renewcommand{\arraystretch}{1.1}
    \resizebox{0.47\textwidth}{!}{
    \begin{tabular}{cclccc}
    \thickhline
    \multirow{2}[1]{*}{Model} & Low-Level & &\multicolumn{3}{c}{High-Level} \\
    \cline{2-2} \cline{4-6} &SSIM\textbf{ $\uparrow$} & & IS\textbf{ $\uparrow$} & IC\textbf{ $\uparrow$} & ACC\textbf{ $\uparrow$} \\
    \hline
    Brain2Image\cite{Brain2Image} & - & & 5.070 & 0.430 & - \\
    EEG-GAN\cite{EEG-GAN} & - & & 5.070 & 0.430 & - \\
    NeuroVision\cite{NeuroVision} & - & & 5.152 & - & - \\
    EEG-VGD\cite{EEG-VGD} & - & & 6.330 & 0.530 & - \\
    DM-RE2I\cite{DM-RE2I} & - & & 12.550 & - & - \\
    DreamDiffusion\cite{DreamDiffusion} & - & & - & - & 45.800 \\
    NeuroImagen\cite{NeuroImagen} & 0.249 & & 33.500 & - & 85.600 \\
    \thickhline
    \end{tabular}%
    }
  \label{tab_resultEEG}%
\end{table}%

\begin{table}[htbp]
  \centering
  \caption{Quantitative Results of AIGC-Brain Models on the \textbf{DNV} Dataset for \textbf{VBV}.}
    \renewcommand{\arraystretch}{1.1}
    \resizebox{0.47\textwidth}{!}{
    \begin{tabular}{cccclcc}
    \thickhline
    \multirow{2}[1]{*}{Model} & \multicolumn{3}{c}{Low-Level} & &\multicolumn{2}{c}{High-Level} \\
    \cline{2-4} \cline{6-7} &SSIM\textbf{ $\uparrow$} &MSE\textbf{ $\downarrow$} &PSNR\textbf{ $\uparrow$} &  & Rank\textbf{ $\downarrow$} & ACC\textbf{ $\uparrow$} \\
    \hline
    DNV\cite{DNV} & 0.16  & 0.090 &- &       & 34.200 & - \\
    f-CVGAN\cite{f-CVGAN} & 0.094 & 0.118 &11.432 &       & -     & - \\
    SSRNM (0.5Hz)\cite{SSRNM} & 0.102 & 0.136 &- &       & 7.180 & - \\
    SSRNM (4Hz)\cite{SSRNM} & 0.086 & 0.128 &- &       & -     & - \\
    Mind-Video\cite{Mind-Video} & -     & 0.171 &- &       & -     & 0.202 \\
    \thickhline
    \end{tabular}%
    }
  \label{tab_resultVid}%
\end{table}%

\subsubsection{Music Metric}
\label{subsubsec: music_metric}
Brain2Music\cite{Brain2Music} evaluates generated music similarity using identification accuracy for two different levels of semantic abstraction and Top-n-class agreement. The latter employs the LEAF classifier\cite{LEAF} on AudioSet classes\cite{AudioSet}, calculating per-class probabilities for original and reconstructed music across genres, instruments, and moods. The top-n agreement measures the overlap between the top-n most probable class labels for the original and reconstructed music in each category. NDMusic\cite{NDMusic} utilizes rank accuracy which refers to the measure of how accurately the music decoding process can identify the specific piece of music each participant was listening to in a trial. It involves assessing the similarity between the original and decoded music for an EEG trial and ranking this similarity against the similarity between the decoded music in that trial and the original music played in all other trials.  

\subsubsection{Text Metric}
\label{subsubsec: text_metric}
The evaluation of decoded text involved several automated low-level and high-level metrics for assessing text similarity. 

\textbf{Low-Level:} Low-level text features provide basic information about the structure and composition of the text but do not capture the deeper meaning or semantic relationships between words, measured by: i){\textbf{Meteor:}} The Meteor metric \cite{Meteor} provides a more robust evaluation by considering not only word overlap but also word order, synonymy, and other linguistic aspects that impact translation quality; ii){\textbf{Rouge:}} Rouge-1 and Rouge-L \cite{Rouge} are specific variants of the Rouge (Recall-Oriented Understudy for Gisting Evaluation) metric that focus on capturing the similarity between the generated and reference summaries; iii) \textbf{BLEU:} BLEU\cite{BLEU} measures the precision of predicted n-grams occurring in the reference sequence; iv) \textbf{WER:} Word Error Rate quantifies the number of edits (insertions, deletions, or substitutions) needed to transform the predicted sequence into the reference sequence.

\textbf{High-Level:} High-level text features are more complex and meaningful representations of text data that capture the context, relationships, and semantics of words and sentences, measured by: i) {\textbf{CLIP:}} A two-way comparison of the output layer of the CLIP-Text \cite{Clip} model; ii) \textbf{BERTScore:} BERTScore \cite{bertscore} utilizes bidirectional transformer language models to represent words and computes a matching score; iii) \textbf{Sentence:} SentenceTransformer similarity score \cite{sentence} is utilized to assess how well the generated text aligns semantically with a reference text.

The text generation metrics for AIGC-Brain models on representative datasets are summarized in Table \ref{tab_resultText}, including AIGC-Brain tasks, datasets, models, and corresponding text metrics. We do not present the performance comparison of the models due to differences in the experimental settings, which would be unfair. For example, despite being under the same dataset, DreamCatcher and BrainCaptioning only use data from Subject-1 and Subject-2 in their experiments, while UniBrain utilizes data from all subjects (1, 2, 5, 7).

\begin{table}[htbp]
  \centering
  \caption{Text Metrics for AIGC-Brain Text Synthesis Models.}
    \renewcommand{\arraystretch}{1.3}
    \resizebox{0.47\textwidth}{!}{
    \begin{tabular}{c|c|c|c}
    \thickhline
    Task  & Dataset & Model & Text Metric \\
    \hline
    \multirow{5}{*}{IBT} & \multirow{3}{*}{NSD} & UniBrain\cite{UniBrain} & Meteor, Rouge, CLIP \\
\cline{3-4}          &       & DreamCatcher\cite{DreamCatcher} & Meteor, Sentence, CLIP \\
\cline{3-4}          &       & BrainCaptioning\cite{BrainCaptioning} & Meteor, Perplexity, Sentence \\
\cline{2-4}          & \multirow{2}{*}{OCD} & GIC-PTL\cite{GIC-PTL} & Rouge \\
\cline{3-4}          &       & GIC-CT\cite{GIC-CT} & Rouge \\
    \hline
    SBT\&VBT & CLSR  & CLSR\cite{CLSR}  & WER, BLEU, Meteor, BERTScore \\
    \thickhline
    \end{tabular}%
    }
  \label{tab_resultText}%
\end{table}%

\section{Conclusions and Prospects}
\label{sec: prospects}
This investigation systematically summarizes the research foundation and recent advancement in the AIGC-Brain domain. On the one hand, the research foundation includes relevant neuroimaging datasets, functional brain regions, and generative models in the mainstream AIGC field. On the other hand, we introduce a taxonomy based on distinctions in model implementation architectures, emphasizing workflows, mapping relationships, and the merits and demerits of each method. This taxonomy serves to delineate the current methodological landscape for AIGC-Brain technologies. Subsequently, we delve into task-specific representative work and implementation details, facilitating a comparative analysis of technological trends, task-specific characteristics, and preferences. To orient researchers to the forefront of the field, we outline cutting-edge work and frameworks for each task.
Additionally, we summarize quality assessment methodologies, covering both qualitative and quantitative dimensions, and showcase performance comparisons on representative datasets for reference. In essence, this survey navigates from the research foundational of AIGC-Brain, through methodology taxonomy, task-specific implementations, and state-of-the-art advancements, to the evaluation of model performance. It furnishes researchers in the field with a comprehensive research continuum, establishing a robust foundation for subsequent technological developments. Moving forward, we discuss the challenges impeding current AIGC-Brain technological advancements and offer insights into potential directions for future developments as follows:

\textbf{Data: }Non-invasive neuroimaging data are generally characterized by low signal-to-noise ratio and high variability, making it difficult to learn and generalize. Moreover, different neuroimaging technologies exhibit preferences for specific types of information, e.g., fMRI emphasizes detailed spatial mapping, EEG prioritizes high temporal resolution, and MEG combines both spatial and temporal information, but neither aspect is prominent. Neuroimaging datasets are basically small in scale due to the complexity and difficulty of neural data acquisition, posing challenges for the application of deep neural networks. Advancements in neuroimaging technology and the availability of larger datasets will contribute to the progress of AIGC-Brain technology.

\textbf{Fidelity: }Unlike the AIGC field, which emphasizes the diversity of generated results, passive brain condition multimodal synthesis content needs to be consistent with the perception received by the brain. That is, generated results in the AIGC-Brain domain are more concerned with detail and semantic fidelity. To this end, future work requires improved cross-modal matching and high-quality generation capabilities in both low-level and high-level aspects.

\textbf{Flexibility: }Flexibility refers to model adaptation and the capability of transferring to different datasets and tasks. For instance, the flexibility of the Map method is significantly greater than that of the BPFA method, which is reflected in: i) The cost of training for the Map method is minimal, whereas BFPA necessitates extensive pretraining and finetuning of the deep generative model; ii) The Map method can flexibly incorporate various prior information or modify the pretrained AIGC decoder to enhance model performance; iii) The Map method is not limited by data variances (e.g., different neuroimaging type and length, etc.), while the BPFA method needs to ensure similarity between upstream pretraining data and downstream data. Overall, Improving model flexibility contributes to model generalization and technology updates.

\textbf{Interpretability:} Decoding brain signals back to perceptual information contributes to unraveling the mechanisms underlying how the brain perceives and comprehends external stimuli. Hence, interpretability is crucial. It is imperative to observe the activation and coordination patterns across various brain regions during the decoding process, as well as dynamic changes. For instance, in the reception of visual information, brain activation regions transition from the low visual cortex (LVC) to the high visual cortex (HVC). Interpretability of AIGC-Brain technologies enhances our understanding of the neural processes involved in decoding sensory information.

\textbf{Real-time: }Given the application of AIGC-Brain technology in BCI systems, real-time processing emerges as a crucial practical consideration. BCI systems necessitate the real-time capabilities of decoding models to achieve online responsiveness and provide instantaneous feedback.

\textbf{Multimodality: }The human brain establishes a One-to-Many relationship between brain modalities and external modalities, as exemplified by brain signals elicited from video stimuli corresponding to video, sound, and text modalities. Furthermore, the perceptual experience in the brain encompasses not only the stimulus itself but also associative information. For instance, upon seeing a photo of a bird, the brain can establish connections with the sound of a bird's call. Therefore, brain signals can concurrently establish connections with multiple external modalities. To the best of our knowledge, among existing AIGC-Brain models, only UniBrain is multimodal, employing a unified AIGC-Brain decoder to simultaneously decode images and text from brain signals elicited by visual stimuli. Looking ahead, we anticipate the emergence of a unified \textbf{\textit{Brain-to-Any}} multimodal synthesis model.

{\small
\bibliographystyle{ieee_fullname}
\bibliography{egbib}
}

\end{document}